\documentclass[twoside]{article}
\usepackage[numbers,square,comma,sort&compress]{natbib}
\usepackage{url}
\usepackage{calc}
\usepackage{amsmath, amssymb, amsthm}
\usepackage{parskip}
\usepackage{color,hyperref}
\usepackage{epsfig}
\usepackage{subfigure}
\usepackage{verbatim}
\usepackage{rotating}
\usepackage{algorithm}
\usepackage{algorithmic}
\usepackage[sort&compress]{natbib}
\usepackage{enumitem}
\usepackage{bbm}
\usepackage{pdflscape}
\usepackage{kky}
\usepackage{hastieDefns}

\newboolean{istwocolumn}
\setboolean{istwocolumn}{false}

\newcommand{\toworkon}[1]{}

\usepackage[accepted]{dupaistats}

\begin{document}



\onecolumn
% \twocolumn[
\aistatstitle{Additive Approximations in High Dimensional 
Nonparametric  Regression via the \salsa}
\aistatsauthor{ Kirthevasan Kandasamy, \hspace{0.01in} Yaoliang Yu  }
\aistatsaddress{ Machine Learning Department, \\
  Carnegie Mellon University, Pittsburgh, PA, USA \\
  \texttt{ \{kandasamy, yaoliang\}@cs.cmu.edu}  }
% ]

\fancyhead[CO]{\small\bf Kirthevasan Kandasamy, \hspace{0.02in} Yaoliang Yu}
\fancyhead[CE]{\small\bf \salsa: Additive Approximations in Regression}

%   \hsize\textwidth
%   \linewidth\hsize \toptitlebar {\centering
%   {\Large\bf \salsas \par}}
%  \bottomtitlebar \vskip 0.2in %plus 1fil minus 0.1in

% For Images
\newcommand{\imarrwtwo}{2.2in}
\newcommand{\imarrwthree}{2.1in}
\newcommand{\imhspthree}{-.1in}
\newcommand{\imarrwthreesmall}{2.2in}
\newcommand{\imhsptwo}{0.3in}
\newcommand{\imarrw}{1.47in} % The width of each subfigure
\newcommand{\imhspace}{-0.25in}
\newcommand{\imtextspace}{-0.2in}
\newcommand{\imcaptionspace}{-0.2in}
\newcommand{\imrowspace}{-0.1in}
\newcommand{\imlabelspace}{-0.0in}
\newcommand{\thmparaspacing}{-0.05in}
\newcommand{\imsinglecol}{2.495in}
\newcommand{\imsinglecolsmall}{2.5in}
\newcommand{\imleftspace}{-0.4in}

\newcommand{\insertFigToy}{
\begin{figure*}
\centering
\hspace{\imleftspace}
\subfigure[]{
  \includegraphics[width=\imarrwthree]{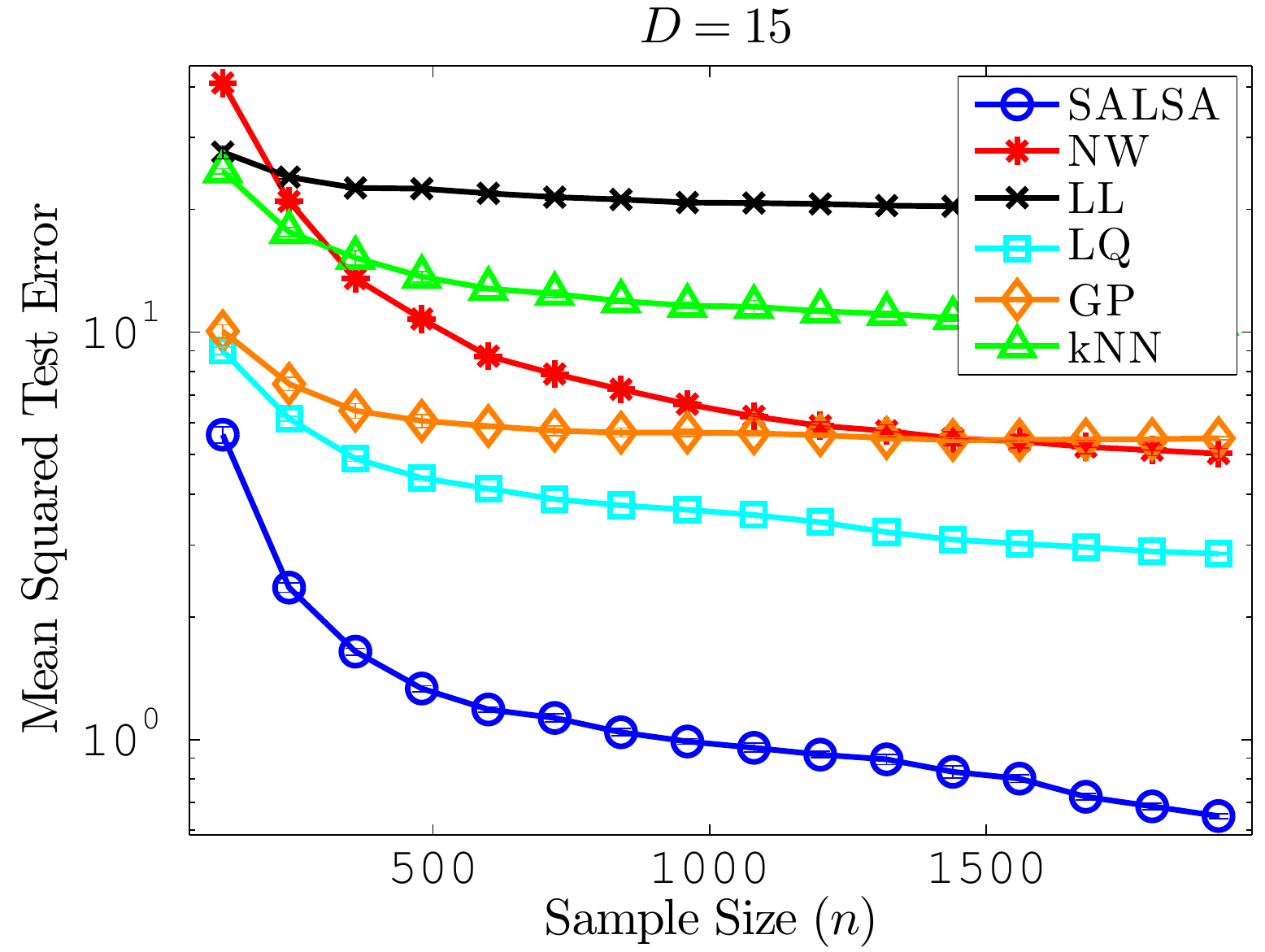} \hspace{\imhspthree}
  \vspace{\imlabelspace}
  \label{fig:knownorder}
}
\subfigure[]{
  \includegraphics[width=\imarrwthree]{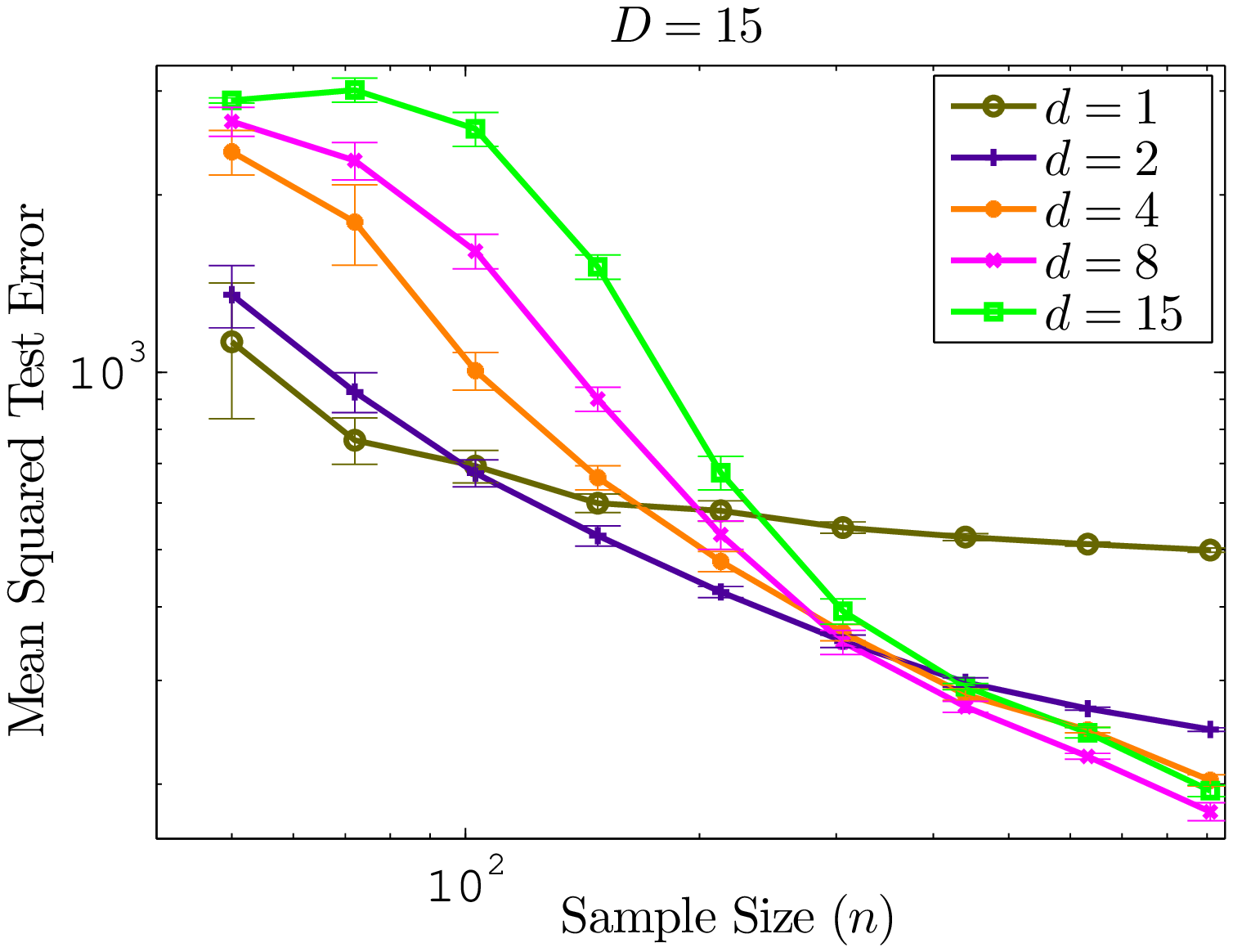} \hspace{\imhspthree}
  \vspace{\imlabelspace}
  \label{fig:comporder}
}
\subfigure[]{
  \includegraphics[width=\imarrwthree]{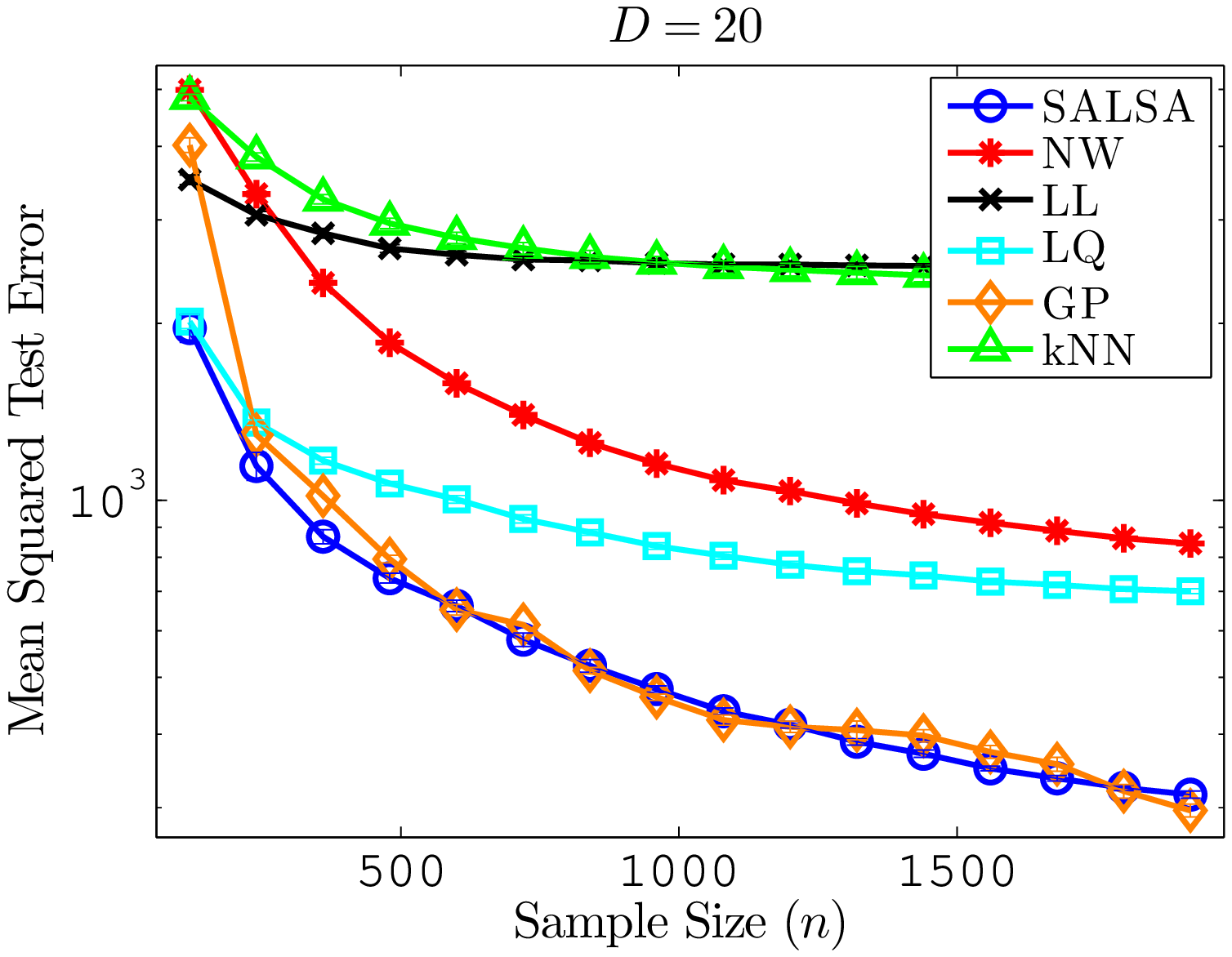} \hspace{\imhspthree}
  \vspace{\imlabelspace}
  \label{fig:cvone}
} \\[-0.1in]
\hspace{\imleftspace}
\subfigure[]{
  \includegraphics[width=\imarrwthree]{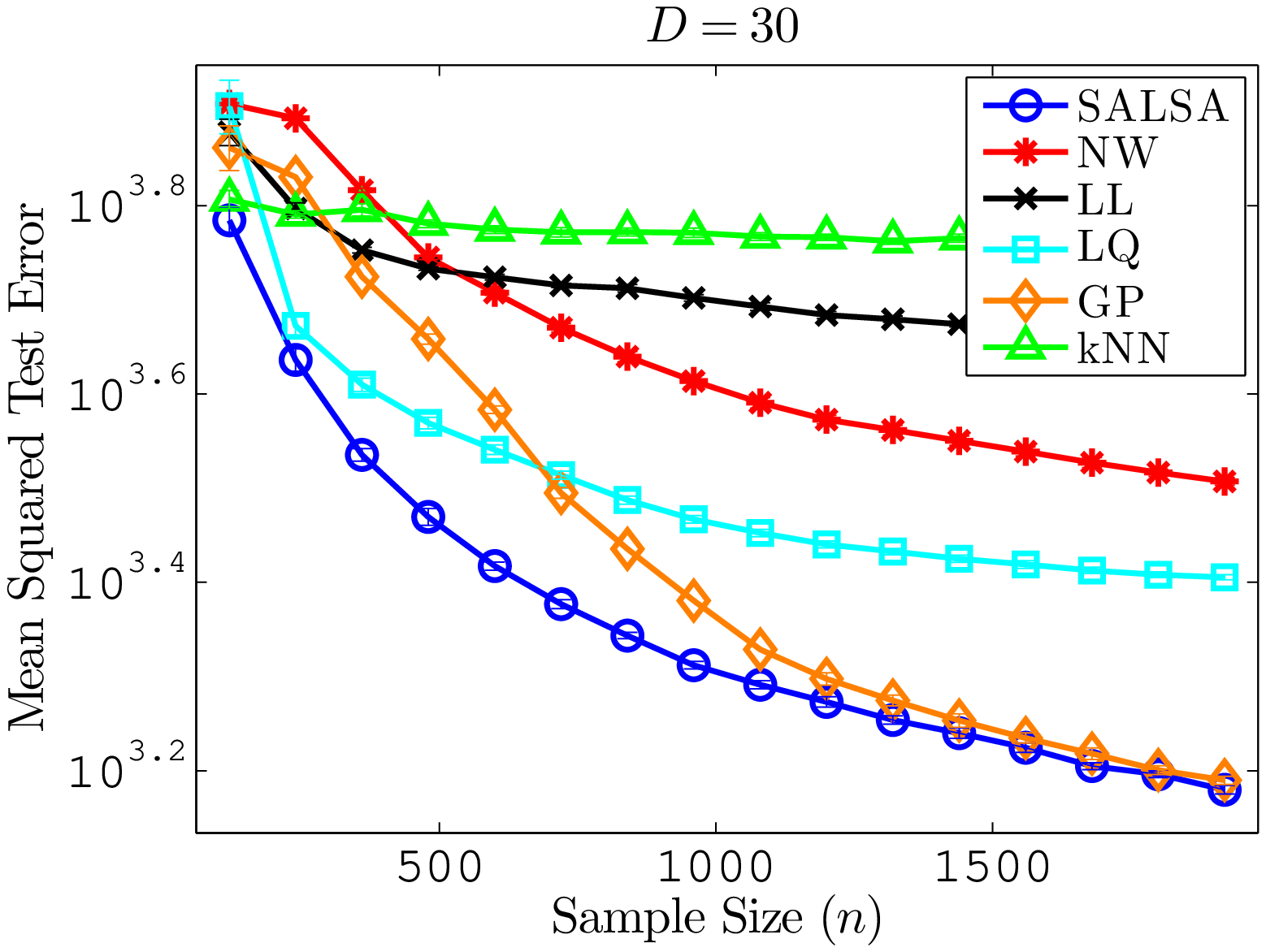} \hspace{\imhspthree}
  \vspace{\imlabelspace}
  \label{fig:cvtwo}
}
\subfigure[]{
  \includegraphics[width=\imarrwthree]{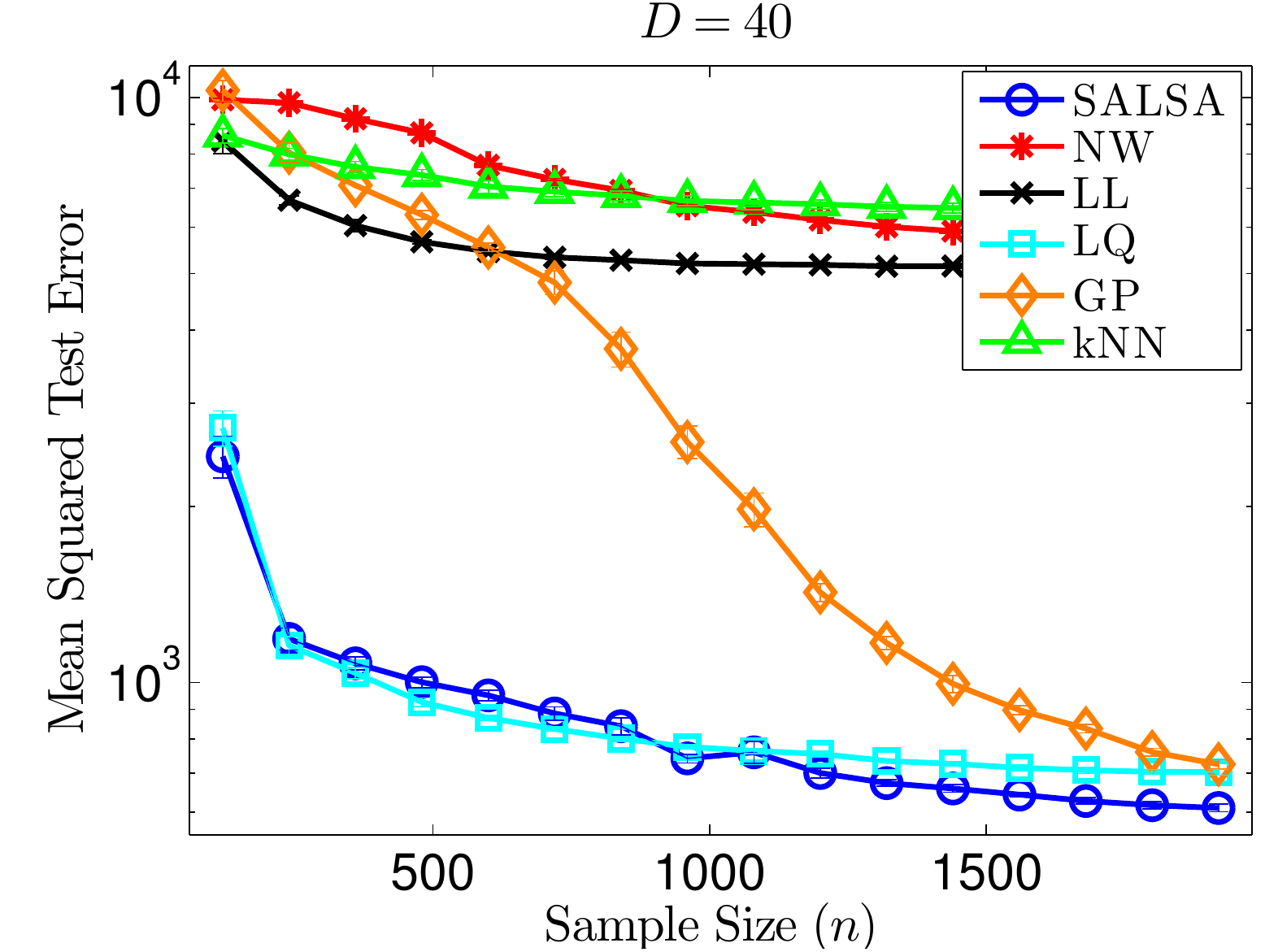} \hspace{\imhspthree}
  \vspace{\imlabelspace}
  \label{fig:cvthree}
}
\subfigure[]{
  \includegraphics[width=\imarrwthree]{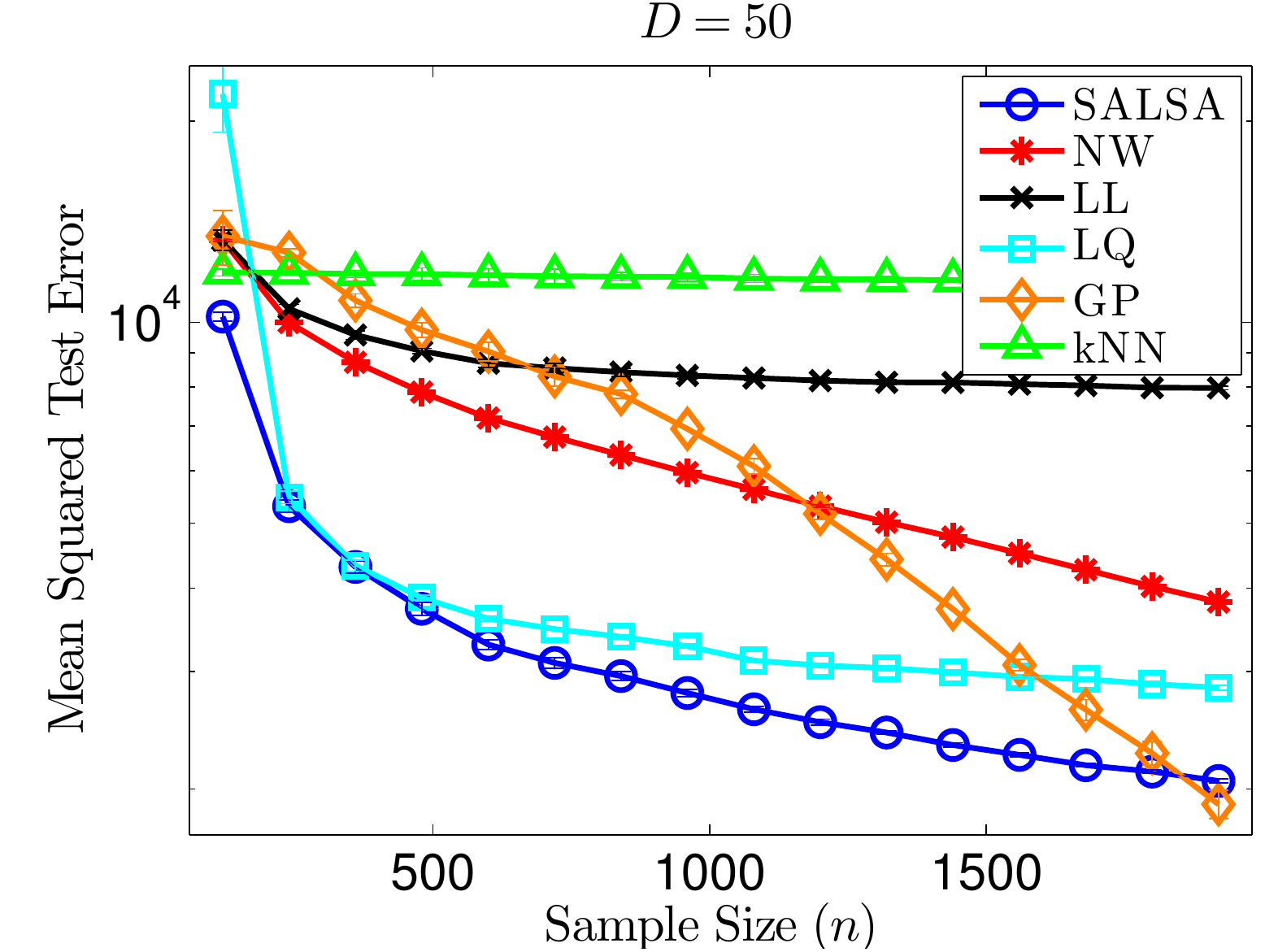} \hspace{\imhspthree}
  \vspace{\imlabelspace}
  \label{fig:cvfour}
}
\vspace{\imcaptionspace}
\caption[]{\small
\subref{fig:knownorder} Comparison of \addkrrs which knows
the additive order of $\regfunc$ against other methods.
\subref{fig:comporder} Comparison of
different choices of $d$ in \salsa. The best $d$ varies with $n$.
\subref{fig:cvone}-\subref{fig:cvfour} Comparison of \addkrrs ($d$ chosen
via cross validation) with alternatives on synthetic datasets.
In all cases, we plot the mean squared prediction error on a
test set of $2000$ points.
All curves are produced by averaging over $10$ trials.
The error bars are not visible in some curves as they are very small.
% \vspace{\imtextspace}
\vspace{-0.1in}
% \vspace{-0.15in}
}
\end{figure*}
}

\newcommand{\insertTableReal}{
\newcommand{\tabletitlenumberspace}{-0.12in}
\begin{sidewaystable}
% \rotatebox{90}{
% \begin{landscape}
% \begin{table}
{ %\footnotesize
\vspace{3.7in}
\centering
% \begin{tabular}{l|c|c|c|c|c|c|c|c|c|c|c|c}
\begin{tabular}{l|c|c|c|c|c|c|c|c|c|c|c}
% Dataset ($D$, $n$) & \addkrr & \krr & \ll & \lq & \svr & \gp & \backf  & \krr & \nw  \\
Dataset ($D$, $n$) & \addkrrs ($d$) & \krr & \knn & \nw & \locallin & \localquad 
  & \sveps &\svnu & \gp & \rt & \gbrt \\
\hline \\[\tabletitlenumberspace]
\hline 
Housing ($12$, $256$)& $\bf{0.26241}$ (9) & $\underline{0.37690}$ & $0.43620$ & $0.38431$ & 
  $\underline{0.31219}$ & \underline{0.35061} &
  $1.15272$ & $0.38600$ & $0.67563$ & $1.06015$ & $0.42951$   \\ \hline
% % Indicate times here
% \emph{Training time} & $2.11 s$ & $0.18 s$ & $0.14 s$ & $0.65 s$ & $0.98 s$ &
% $1.54 s$ & $2.16 $& $0.20 s$ & $2.99 s$ & $1.67 s$  & $0.15 s$
% \\ \hline
% 
Galaxy ($20$,$2000$)
& $\bf{0.00014}$ (4) & $0.01317$ & $0.25854$ & $0.30615$ & $0.01676$ &
\underline{$0.00175$} &
 $0.65280$ & $0.15798$ & $0.00221$ & $0.02293$ & $0.01405$  \\ \hline
fMRI ($100$,$700$)
& $\bf{0.80730}$ (2) & $0.86495$ & $0.85645$ & $0.84989$ & $0.91098$ & $1.14079$ &
$\underline{0.81080}$ & $\underline{0.81376}$ & $0.84766$ & $1.52834$  & $0.87326$ \\ \hline
Insulin ($50$,$256$)
& $\bf{1.02062}$ (3) & $\underline{1.09023}$ & $1.15578$ & $1.18070$ & 
$\underline{1.06457}$ & $1.35747$ &
$1.10725$ & $1.09140$ & $1.22404$ & $1.58009$ & \underline{$1.06624$} \\ \hline
Skillcraft ($18$,$1700$)
& $\bf{0.54695}$ (1) & $\underline{0.54803}$ & $0.67155$ & $0.73258$ & $0.60581$ & $1.29690$ &
$0.71261$ &
$0.66311$ & $\underline{0.54816}$ & $1.08047$ & $0.57273$  \\ \hline
% 
% Blog ($91$,$700$)
% & $\underline{0.03053}$ (2) & $0.47249$ & $4.21784$ & $0.56196$ & $0.67570$ & $0.79760$ & 
% $0.75840$ &
% $0.55447$ & $0.65800$ & $2.99038$ & $1.92752$  \\ \hline
% 
School ($36$,$90$)
& $\underline{1.32008}$ (2) & $1.64315$ & $\underline{1.37910}$ & $\bf{1.14866}$ & 
$2.13390$ & $4.79447$ &
$\underline{1.38173}$ & $1.48746$ & $1.64215$ & $1.99244$ & $2.09863$ \\ \hline
CCPP* ($59$,$2000$)
& $\underline{0.06782}$ (2) & $0.08038$ & $0.32017$ & $0.33863$ & 
\underline{$0.07568$} & $\underline{0.06779}$ &
$0.33707$ & $0.094493$ & $0.11128$ & $1.04527$ & \underline{$0.06181$} \\ \hline
% 
% Dataset ($D$, $n$) & \addkrr & \krr & \knn & \nw & \locallin & \localquad 
%   &\localcubic & \sveps &\svnu & \gp & \rt & \rbfi \\
% 
Bleeding ($100$,$200$)
& $\underline{0.00123}$ (5) & $0.10633$ & $0.16727$ & $0.19284$ & $\bf{2.48\textrm{e}
-6}$ & $0.00614$ &
$0.36505$ & $0.03168$ & $\underline{6.68\textrm{e}-6}$ & $0.18292$ & $0.19076$ \\ \hline
Speech ($21$, $520$)& $\underline{0.02246}$ (2) & $0.03036$ & $0.09348$ & $0.11207$ & 
$0.03373$ & $\underline{0.02404}$ & 
$\underline{0.22431}$ & $0.06994$ & $0.02531$ & $0.05430$ & $0.03515$ \\
% % Indicate times here
\hspace{0.3in}\emph{Training time} & $4.71 s$ & $0.8 s$ & $0.18 s$ & $1.81 s$ & $4.53 s$ &
$6.80 s$ & $0.24 s$ & $27.43 s$ & $6.34 s$  & $0.21 s$ & $5.30 s$
\\ \hline
Music ($90$,$1000$)
& $\underline{0.62512}$ (3) & $\underline{0.61244}$ & $0.71141$ & $0.75225$ & $0.67271$ & $1.31957$ &
$0.75420$ & $\bf{0.59399}$ & $\underline{0.62429}$ & $1.45983$ & $0.66652$  \\ \hline
Telemonit ($19$, $1000$)
& $\underline{0.03473}$ (9) & $0.05640$ & $0.09262$ & $0.21198$ & $0.08253$ & $0.18399$ &
$0.33902$ & $0.05246$ & $0.03948$ & $\bf{0.01375}$ & $0.04371$ \\ \hline
Propulsion ($15$,$200$) & $\underline{0.00881}$ (8) & $0.05010$ & $0.14614$ & $0.11237$ &
$1.12712$ &
$1.12801$ & $0.74511$ & $\underline{0.00910}$ & $\underline{0.00355}$ & $0.02341$ &
$\bf{0.00061}$   \\ \hline
Airfoil* ($40$,$750$)
& $\underline{0.51756}$ (5) & $0.53111$ & $0.85879$ & $0.86752$ & $0.51877$ & 
$\underline{0.51105}$ &
$0.64853$ & $0.55118$ & $0.54494$ & $\underline{0.45249}$ & $\bf{0.34461}$  \\ \hline
Forestfires($10$, $211$)
& $0.35301$ (3) & $\underline{0.28771}$ & $0.36571$ & $0.37199$ 
& $0.35462$ & $12.5727$ &
$0.70154$ & $0.43142$ & $\underline{0.29038}$ & $0.41531$ & \underline{$0.26162$}  \\ \hline
Brain ($29$,$300$)
& $0.00036$ (2) & $0.01239$ & $0.24429$ & $0.22929$ & $1.23412$ & $2.45781$ &
$0.27204$ & $0.05556$ & $\underline{5\textrm{e}-14}$ & $0.00796$  & $0.00693$ \\ \hline 
\hline \\[-0.06in]
% 
% Dataset ($D$, $n$) & \addkrr & \krr & \knn & \nw & \locallin & \localquad 
%   &\localcubic & \sveps &\svnu & \gp & \rt & \rbfi \\
% \hline \\[-0.00in]
% \hline
  & \rbfi
 & \mfp & \si & \backf & \mars & \cosso & \spam &\addgp
  & \lr & \lasso & \lar  \\
\hline \\[\tabletitlenumberspace]
\hline 
Housing ($12$, $256$)& $0.64871$ & $\underline{0.38256}$ & $0.50445$ & $0.64218$ & $0.42379$ & $1.30965$ &
$0.81653$ & $0.45656$ & $95.60708$ & $0.44515$ & $0.84410$  \\ \hline
% % Indicate times here
% \emph{Training time} & $0.04 s$ & $0.49 s$ & $0.12 s$ & $6.30 s$ & $0.68 s$ &
% $1.74 s$ & $17.74 s$& $\sim 3\, mins$ & $0.01 s$ & $1.08 s$  & $0.06 s$
% \\ \hline
% 
Galaxy ($20$,$2000$) & $0.01532$
& $\underline{0.00116}$ & $0.92189$ & $0.94165$ & $\underline{0.00163}$ 
& $\underline{0.00153}$ & $0.95415$ 
& $-$ & $0.13902$ & $0.02392$ & $1.02315$  \\ \hline
fMRI ($100$,$700$)
& $1.38585$ & $1.42795$ & $0.90595$ & $0.86197$ & $0.90850$ & $0.82448$ & $0.88014$ &
$-$&
$\underline{0.81005}$ & $\underline{0.81390}$ & $0.88351$  \\ \hline
Insulin ($50$,$256$)
& $1.22404$ 
& $1.78252$ & $1.20771$ & $1.16524$ & $1.10359$ & $1.13791$ & $1.20345$ & $-$
& $\bf{1.02051}$ &
$1.11034$ & $1.22404$  \\ \hline
Skillcraft ($18$,$1700$)
& $0.81966$ & $1.07195$ & $0.87677$ & $0.83733$ & $\bf{0.54595}$ &
$\underline{0.55514}$ & $0.905445$ &
$-$ & $0.70910$ & $0.66496$ & $1.00048$  \\ \hline
% 
% Blog ($50$,$700$)
% & $4.32056$ & $0.27582$ & $\underline{0.05404}$ & $\underline{0.03497}$ & $0.52079$ &
% $0.53500$ & $\bf{0.02882}$ & $-$ &
% $0.35164$ & $0.30992$ & $\underline{0.03060}$  \\ \hline
% 
School ($36$,$90$)
& $1.61927$ & $1.72657$ & $1.52374$ & $1.48866$ & $1.44453$ & $1.48046$ & $1.59328$ &
$-$ & $1.53416$ & $\underline{1.34467}$ & $1.64330$  \\ \hline
CCPP* ($59$,$2000$)
& $1.04257$ & $0.10513$ & $0.97223$ & $0.88084$ & $0.08189$ & $0.96844$ & $\bf{0.06469}$ &
$-$ & $0.07641$ & 
$0.07395$ & $1.04527$  \\ \hline
Bleeding ($100$,$200$)
& $0.13872$ & $0.00210$ & $0.24918$ & $0.37840$ & $0.00497$ & $0.31362$ & $0.41735$ &
 $-$ & $\underline{0.00001}$ & $\underline{0.00191}$ & $0.43488$  \\ \hline
Speech ($21$, $520$)
& $0.03339$  
& $0.02843$ & $0.39883$ & $0.36793$ & $\bf{0.01647}$ & $0.34863$ &
$0.66009$ & $\underline{0.02310}$ & $0.07392$ & $0.07303$ & $0.73916$  \\
% Indicate times here
 \hspace{0.3in}\emph{Training time} & $0.12 s$ & $5.70 s$ & $0.66 s$ & $27.93 s$ & $8.11 s$ &
$9.40 s$ & $76.39 s$& $79\,mins$ & $0.03 s$ & $15.94 s$  & $0.06 s$
\\ \hline
Music ($90$,$1000$)
& $0.78482$ 
& $1.28709$ & $0.77347$ & $0.75646$ & $0.88779$ &  $0.79816$ & $0.76830$ & $-$ & 
$0.67777$ & $\underline{0.63486}$ & $0.78533$  \\ \hline
Telemonit ($19$, $1000$)
& $\underline{0.02872}$  
& $\underline{0.01491}$ & $0.65386$ & $0.84412$ & $\underline{0.02400}$ & $5.71918$ & $0.86425$ & 
$-$ & $0.08053$ &
$0.08629$ & $0.94943$  \\ \hline
Propulsion ($15$,$200$) 
& $0.05832$ 
& $0.02341$ & $0.27768$ & $0.56418$ & $0.0129$ & $\underline{0.00094}$ & $1.11210$ &
$0.01435$ & $0.01490$ & $0.02481$ & $10.2341$  \\ \hline
Airfoil* ($40$,$750$)
& $1.00909$ & $\underline{0.46714}$ & $0.99413$ & $0.96668$ & $0.54552$ & $0.51782$ & 
$0.96231$ & $-$ & $0.53187$ & $0.51986$ & $1.00910$  \\ \hline
Forestfires($10$, $211$)
& $0.45773$ 
& $0.47749$ & $0.78057$ & $0.88979$ & $\underline{0.33891}$ & $0.37534$ & $0.96944$ 
& $\bf{0.17024}$ &
$0.47892$ & $0.51934$ & $1.05415$  \\ \hline
Brain ($29$,$300$)
& $0.97815$ & $\bf{2\textrm{e}-37}$ & $0.81711$ & $0.63700$ &
$\underline{5\textrm{e}-31}$ & $\underline{2\textrm{e}-6}$ & $0.89533$ &
$-$ &$\underline{4\textrm{e}-13}$& $0.00089$ & $1.04216$  \\ \hline
% 
% rbfi  & \mfp & \si & \backf & \mars & \cosso & \spam &\addgp
%   & \lr & \lasso & \lar & &  \\
% 
\hline
\hline
\end{tabular}
\vspace{-0.1in}
\caption{ \small
The average squared error on the test set for all methods on 16 datasets.
The dimensionality and sample size $n$ are indicated next to the dataset.
The best method(s) for each dataset are in bold. The second to fifth methods 
are underlined. 
% For the Speech dataset we have indicated the training time for each method.
For \addkrrs we have also indicated the order $d$ chosen by our cross validation
procedure in parantheses.
The datasets with a * are actually lower dimensional datasets from the UCI repository. 
But we
artificially increase the dimensionality by inserting random values for the remaining
coordinates. Even though, this doesn't change the function value it makes the
regression problem harder.
}
\label{tb:realData}
}
% \end{table}
\end{sidewaystable}
% \end{landscape}
% }
}

\begin{abstract}
\vspace{0.05in}
High dimensional nonparametric regression is an inherently difficult problem with 
known lower bounds depending exponentially in dimension. 
A popular strategy to alleviate this curse of dimensionality  
has been to use additive models of \emph{first order}, which model the regression
function as a sum of independent functions on each dimension. 
Though useful in controlling the variance of the estimate, such models are
often too restrictive in practical settings.
Between non-additive models which often have large variance and first order
additive models which have large bias, there has been little work to
exploit the trade-off in the middle via additive models of intermediate order.
In this work, we propose \salsa, which bridges this gap by allowing
interactions between variables, but controls model capacity by limiting the order of
interactions. 
\salsas minimises the residual sum of squares with squared RKHS norm penalties.
Algorithmically, it can be viewed as Kernel Ridge Regression with an additive kernel.
When the regression function is additive, the excess risk is only polynomial 
in dimension. 
Using the Girard-Newton formulae, we efficiently sum over a combinatorial number
of terms in the additive expansion.
Via a comparison on $15$ real datasets, we show that our method is competitive
against $21$ other alternatives.
\end{abstract}

% Our assumptions are more expressive than
% existing literature on additive models for regression.

% We prove that when the regression function is additive, 
% the sample complexity has only polynomial 
% dependence on dimension. 

% We propose a computationally efficient procedure to sum over a combinatorial number
% of terms in the additive expansion.
% We propose an efficient procedure to solve a seemingly combinatorially
% difficult computational problem 

%!TEX root = v2hastie.tex

\section{Introduction}
\label{sec:intro}

Given \iid samples $(X_i,Y_i)_{i=1}^n$ from some distribution
$\PXY$, on $\Xcal\times \Ycal \subset \RR^D \times \RR$, 
the goal of least squares regression is to estimate the
regression function $\regfunc(x) = \EE[Y|X=x]$.
A popular approach is linear regression which models $\regfunc$ as a
linear combination of the variables $x$, i.e. $f(x) = \beta^\top x$ for some $\beta \in
\RR^D$. Linear Regression is typically solved by minimising the sum of squared errors on
the training set subject to a complexity penalty on $\beta$.
Such \emph{parametric} methods are conceptually  simple and have desirable
statistical properties when the problem meets the assumption. However, the
parametric assumption is generally too restrictive for many real problems.

Nonparametric regression refers to a suite of methods that typically
only assume smoothness on $\regfunc$. 
They present a more compelling framework for regression since
they encompass a richer class of functions than parametric models do.
However they suffer from severe drawbacks in high dimensional settings. 
The excess risk of nonparametric methods has exponential dependence on
dimension. Current lower bounds~\cite{gyorfi02distributionfree} suggest that
this dependence is unavoidable.
Therefore, to make progress stronger assumptions on $\regfunc$
beyond just smoothness are necessary.
In this light,
a common simplification  has been to assume that $\regfunc$ decomposes into the additive
form $\regfunc(x) = \regfuncii{1}(x_1) + \regfuncii{2}(x_2) 
+ \dots + \regfuncii{D}(x_D)$ \cite{hastie90gam,lafferty05rodeo,ravikumar09spam}. 
In this exposition, we refer to such models as \emph{first order}
additive models. Under this assumption, the excess risk improves significantly.

That said, the first order assumption is often too biased in practice since it
ignores interactions between variables. It is natural to ask if we could
consider additive models which permit interactions. For instance, a
second order model has the expansion 
$\regfunc(x) = \regfuncii{1}(x_1,x_2)
+ \regfuncii{2}(x_1,x_3) + \dots\;$. In general, we may consider $d$ orders of
interaction which have ${D \choose d}$ terms in the expansion.
If $d \ll D$, we may allow for a richer class of functions than first order
models, and hopefully still be able to control the excess risk.

% When $\regfunc$ exhibits additive structure, using an additive
% model for estimation is understandably reasonable. However, 
Even when $\regfunc$ is
not additive, using an additive approximation has its advantages. 
It is a well understood statistical concept that when we only have few samples, using a
simpler model to fit our data gives us a better trade-off for variance against bias. 
Since additive models are \emph{statistically simpler} they  may give us better 
estimates due to reduced variance.
In most nonparametric regression methods, the bias-variance trade-off
is managed via a parameter such as the bandwidth of a kernel or a 
complexity penalty.
In this work, we demonstrate that this trade-off can also be controlled
via additive models with different orders of
  interaction. 
Intuitively, we might use low order interactions with few data
points but with more data we can increase model capacity via higher 
order interactions. Indeed, our experiments substantiate this intuition:
additive models do well on several datasets in which $\regfunc$ is not 
necessarily additive.
% Even though our theoretical analysis assumes $\regfunc$ is additive,

There are \textbf{two key messages in this paper}. 
The first is that we should use additive models
in high dimensional regression to reduce the variance of the estimate.
The second is that it is necessary 
to model beyond just first order models to
reduce the bias.
Our contributions in this paper are:
\begin{enumerate}[leftmargin=*]
\item 
We formulate additive models for nonparametric regression beyond first order models.
Our method \salsas--for \emph{Shrunk Additive Least Squares Approximation}--
estimates a $d$\superscript{th} order additive function containing
$D \choose d$ terms in its expansion.
Despite this, 
the computational complexity of \salsas is $\bigO(Dd^2)$.
% minimises the sum of squared errors subject to complexity penalties on
% each of the $D \choose d$ terms in the additive expansion.
% (shrunk as in shrinkage due to regularisation).
% Our method, which optimises for the estimate over a sum RKHS, is called
% \salsas--for \emph{Shrunk Additive Least Squares Approximations}
% (shrunk as in shrinkage due to regularisation).
% Though there are $D \choose d$ terms in the additive expansion, 
% the computational complexity of \salsas is $\bigO(Dd^2)$.
\vspace{\itemspace}
\item Our theoretical analysis bounds the excess risk for \addkrrs for 
(i) additive $\regfunc$ under reproducing kernel Hilbert space assumptions and 
(ii) non-additive $\regfunc$ in the agnostic setting.
In (i), the excess risk has only polynomial dependence on $D$.
\vspace{\itemspace}
\item We compare our method against $21$ alternatives
on synthetic and  $15$ real datasets. 
\addkrrs is more consistent and in many cases outperforms other methods.
Our software and datasets are available at 
{\small\texttt{github.com/kirthevasank/salsa}}.
Our implementation of locally polynomial regression is also released as part of
this paper and is made available at
{\small\texttt{github.com/kirthevasank/local-poly-reg}}.
\end{enumerate}

Before we proceed we make an essential observation.
When parametric assumptions are true, parametric regression methods can scale 
both statistically and
computationally to possibly several thousands of dimensions. 
However, it is common knowledge in the statistics community that nonparametric 
regression can be reliably applied only in very low dimensions 
with reasonable data set sizes. %as the excess risk depends exponentially on $D$. 
Even $D=10$ is considered ``high" for nonparametric methods.
In this work we aim to statistically scale nonparametric 
regression to dimensions on the order $10\textrm{--}100$ while addressing the
computational challenges in doing so.
% \toworkon{this is in response to the NIPS reviewers. Is this explanation
% too long ?}

\subsection*{Related Work}
% \vspace{-0.05in}
\label{sec:relwork}

A plurality  of work in high dimensional regression focuses on first order additive
models. One of the most popular techniques is the back-fitting
algorithm~\cite{hastie01statisticallearning} which iteratively approximates
$\regfunc$ via a sum of $D$ one dimensional functions.
Some variants such as RODEO~\cite{lafferty05rodeo} and SpAM~\cite{ravikumar09spam}
 study first order models
in variable selection/sparsity settings.
MARS \cite{friedman91mars} 
uses a sum of splines on individual dimensions but allows interactions between
variables via products of hinge functions at selected knot points.
\citet{lou13intelligible} model $\regfunc$ as a first order model plus a
sparse collection of pairwise interactions. However, restricting ourselves to only 
to a sparse collection of second order interactions  might
be too biased in practice.
COSSO~\cite{lin2006cosso} study higher order models but when you need only a sparse
collection of them.
In Section~\ref{sec:experiments} we list several other parametric and nonparametric
methods used in regression.

Our approach is based on additive kernels and builds on Kernel Ridge
Regression~\cite{steinwart08svms,zhang05learning}.
Using additive kernels to encode and identify structure in the problem is fairly
common in Machine Learning literature. A large line of work, in what has to come to
be known as Multiple Kernel Learning (MKL), focuses on precisely this
problem~\cite{gonen11mkl,xu10kernelgrouplasso,bach08grouplasso}.
Additive models have also been studied in Gaussian process
literature via additive kernels~\cite{duvenaud11additivegps,plate99additivegps}.
% In fact, we use some ideas from~\citet{duvenaud11additivegps} in this work.
However, they treat the additive model just as a heuristic whereas we also
provide a theoretical analysis of our methods.
% Further marginal likelihood maximisation in Additive GPs to fit the parameters can
% be very expensive as we found in our experiments.

\section{Preliminaries}
\label{sec:setup}

% We begin with a brief review on Reproducing Kernel Hilbert Spaces and Kernel Ridge
% Regresion (\krr) to facilitate the ensuing discussion.
We begin with a brief review of some background material.
% to facilitate the ensuing discussion.
We are given \iid data $\XYn$ sampled from some distribution $\PXY$ on a 
compact space $\Xcal\times \Ycal \subset \RR^D \times \RR$.
Let the marginal distribution of $X$ on $\Xcal$ be $\px$ and the $L_2(\PX)$ norm
be $\|f\|_2^2 = \int f^2\ud\PX$.
We wish to use the data
to  find a function $\func:\Xcal
\rightarrow \RR$ with small risk
\[
\Rcal(f) = \int_{\Xcal\times \Ycal} (y - f(x))^2 \ud \pxy(x,y)
 = \EE[(Y- f(X))^2].
\]
% Here, the expectation in the first step is over the data and the second is over both
% the data and $\PX$.
It is well known that $\Rcal$ is minimised by the regression function
$\regfunc(\cdot) = \EE_{XY}[Y|X=\cdot]$ and the \emph{excess risk} 
for any $f$ is 
$\Rcal(f) - R(\regfunc) = \|f-\regfunc\|^2_2$
\cite{gyorfi02distributionfree}.
Our goal is to develop an estimate that has low expected excess risk
$\EE\Rcal(\funchat) - \Rcal(\regfunc) = \EE[\|\funchat-\regfunc\|^2_2]$, where the 
expectation is taken with respect to realisations of the data $\XYn$.
% % % For simplicity of this exposition we will assume $\Xcal = [0,1]^D$ and that 
% % % 
% % % Recall that our goal in nonparametric regression
% % % is to estimate the regression function $\func:\Xcal \rightarrow \RR$ where 
% % % $\func(x) = \EE[Y|X=x]$. For simplicity we will assume 
% % % $\Xcal$ is a compact subset of $\RR^D$. 

Some smoothness conditions on $\regfunc$ are required to make regression
tractable. A common assumption is that $\regfunc$ has bounded norm
in the reproducing
kernel Hilbert space (RKHS) $\Hcalk$ of a continuous positive definite kernel 
$\kernel:\Xcal\times\Xcal \rightarrow \RR$.
By Mercer's theorem \cite{scholkopf01kernels}, $\kernel$ permits an
eigenexpansion of the form $\kernel(x,x') = \sum_{j=1}^\infty \mu_j \phi_j(x)
\phi_j(x')$ where $\mu_1\geq\mu_2\geq \dots \geq 0$ are the eigenvalues 
of the expansion 
and $\phi_1,\phi_2, \dots$ are an orthonormal basis for $\Ltwo(\PX)$. 
% The infinite sum converges uniformly and absolutely.

Kernel Ridge Regression (\krr) is a popular technique for nonparametric regression.
It is characterised as the solution of the following optimisation problem
over the RKHS of some kernel $\kernel$.  % added to long version
\begin{align}
\funchat = \argmin_{f \in \Hcalk} \lambda \|f\|^2_\Hcalk + 
  \frac{1}{n} \sum_{i=1}^n (Y_i - f(X_i))^2.
\label{eqn:krrDefn}
\end{align}
Here $\lambda$ is the regularisation coefficient to control the variance of the
estimate and is decreasing with more data.
Via the representer theorem \cite{scholkopf01kernels,steinwart08svms}, 
we know that the solution lies
in the linear span of the canonical maps of the training points $\Xn$ --
i.e. $\funchat(\cdot) = \sum_i \alpha_i \kernel(\cdot, X_i)$. This reduces the above
objective to $\alphahat = \argmin_{\alpha\in\RR^n} \lambda \alpha^\top K \alpha 
+ \tfrac{1}{n}\|Y - K\alpha\|_2^2$ where $K \in \RR^{n\times n}$ is the kernel matrix with
$K_{ij} = \kernel(X_i,X_j)$. The problem has the closed form solution $\alphahat =
(K + \lambda n I)^{-1} Y$.
\krrs has been analysed extensively under different
assumptions on $\regfunc$; see \cite{steinwart09optimal,zhang05learning,
steinwart08svms} and references therein.
Unfortunately, as is the case with many nonparametric methods, \krrs suffers from the
curse of dimensionality as its excess risk is exponential in $D$.

% \subsection*{Assumptions}
\textbf{Additive assumption:} 
To make progress in high dimensions, we
assume that $\regfunc$ decomposes into the following additive form that contains 
interactions of $d$ orders among the variables. 
(Later on, we will analyse non-additive $\regfunc$.)
\ifthenelse{\boolean{istwocolumn}}{
\begin{align*}
\regfunc(x) &= %\sum_{j=1}^{\Md} \regfuncj(\xj)  %\\
%   &= 
  \sum_{1\leq i_1<i_2<\dots<i_d\leq D} 
  \regfuncj(x_{i_1},x_{i_2},\dots, x_{i_d}),
\numberthis \label{eqn:addAssumption}
\end{align*}
We will write, $\regfunc(x) = \sum_{j=1}^{\Md} \regfuncj(\xj)$ where $\Md = {D \choose d}$,
and $\xj$ denotes the subset $(x_{i_1},x_{i_2},\dots,x_{i_d})$.
} {
Writing $\Md = {D \choose d}$,
\begin{align}
\regfunc(x) = \sum_{j=1}^{\Md} \regfuncj(\xj) =  
\sum_{1\leq i_1<i_2<\dots<i_d\leq D} 
  \regfuncj(x_{i_1},x_{i_2},\dots, x_{i_d}),
% \regfunc(x) = \sum_{\jb} \regfuncj(\xj) =  
% \sum_{1\leq j_1<j_2<\dots<j_d\leq D} 
%   \regfuncj(x_{j_1},x_{j_2},\dots, x_{j_d})
\label{eqn:addAssumption}
\end{align}
where $\xj$ denotes the subset $(x_{i_1},x_{i_2},\dots,x_{i_d})$.
}
We are primarily interested in the setting $d \ll D$.
While there are a large number of $\regfuncj$'s, each of them only permits
interactions of at most $d$ variables. We will show that this assumption does in fact
reduce the statistical complexity of the function to be estimated.
The first order additive assumption is equivalent to setting $d=1$ above.
A potential difficulty with the above assumption is the combinatorial computational
cost in estimating all $\regfuncj$'s when $d>1$.
% We feel that this has been a critical bottleneck in extending additive 
% models beyond first order models.
We circumvent this bottleneck using two strategems: a classical result
from RKHS theory, and 
a computational trick using elementary symmetric
polynomials used before by~\citet{shawe2004kernel,duvenaud11additivegps}
in the kernel literature for additive kernels.

%!TEX root = v2hastie.tex

% \section{Shrunk Additive Least Squares Approximation}
% \section{\normalsize{Shrunk Additive Least Squares Approximation}}
\section{\salsa}
\label{sec:addkrr}

To extend \krrs to additive models we first
define kernels $\kernelj$ that act on each subset $\xj$. 
We then optimise the following objective jointly over $\funchatj \in \Hcalkj, j =
1\dots,\Md$. 
\ifthenelse{\boolean{istwocolumn}}{
% \begingroup
\begin{align*}
% \allowdisplaybreaks
& \{\funchatj\}_{j=1}^{\Md} =
\argmin_{\funcj \in \Hcalkj, j = 1,\dots, \Md} 
%   F_\lambda\big(\{\funchatj\}_{j=1}^{\Md}\big) 
%   \hspace{0.1in}\textrm{where, } \\
\;\;\lambda \, \sum_{j=1}^\Md \|\funcj\|^2_{\Hcalkj} \;+ \\
&\hspace{1.0in}
  \frac{1}{n}\sum_{i=1}^n \Big(Y_i - \sum_{j=1}^{\Md} \funcj (\Xj_i) \Big)^2.
\numberthis \label{eqn:addObj}
\end{align*}
% \endgroup
}{
\begin{align*}
& \{\funchatj\}_{j=1}^{\Md} =
\argmin_{\funcj \in \Hcalkj, j = 1,\dots, M} 
 \lambda \, \sum_{j=1}^\Md \|\funcj\|^2_{\Hcalkj} +
  \frac{1}{n}\sum_{i=1}^n \Big(Y_i - \sum_{j=1}^{\Md} \funcj (\Xj_i) \Big)^2.
\numberthis \label{eqn:addObj}
\end{align*}
}Our estimate for $\func$ is then $\funchat(\cdot) = \sum_j \funchatj(\cdot)$.
At first, this appears troublesome since it requres optimising over $n\Md$ parameters
$(\alpha^{(j)}_i), j = 1,\dots,\Md, i = 1,\dots,n$. 
However, from the work of~\citet{aronszajn50rkhs}, we know that the solution 
of~\eqref{eqn:addObj} lies in the RKHS of the sum kernel $k$ 
\ifthenelse{\boolean{istwocolumn}}{
\begin{align*}
&\kerneld(x,x') = \sum_{j=1}^{M_d} \kernelj(\xj,{\xj}') 
\numberthis \label{eqn:sumkernel} \\
&\hspace{0.15in}= \sum_{1\leq i_1<\dots<i_d\leq D} 
  \kernelj([x_{i_1},\dots, x_{i_d}], [x'_{i_1},\dots, x'_{i_d}]).
\end{align*}
}{
\begin{equation*}
\kerneld(x,x') = \sum_{j=1}^{M_d} \kernelj(\xj,{\xj}')
= \sum_{1\leq i_1<i_2<\dots<i_d\leq D} 
  \kernelj([x_{i_1},\dots, x_{i_d}], [x'_{i_1},\dots, x'_{i_d}]).
\numberthis
\label{eqn:sumkernel}
\end{equation*}
}See Remark~\ref{rmk:sumRKHS} in Appendix~\ref{sec:appMainThm} for a proof.
Hence, the solution $\funchat$ can be written in  the form 
$\funchat(\cdot) =  \sum_i \alpha_i \kerneld(\cdot, X_i)$
This is convenient since we only need to optimise over $n$
parameters despite the combinatorial number of kernels.
Moreover, it is straightforward to see that the solution is obtained by
solving~\eqref{eqn:krrDefn} by plugging in the sum kernel $\kerneld$ for $\kernel$.
Consequently $\funchatj = \sum_i\alphahat_i \kernelj(\cdot,\Xj_i)$
and $\funchat = \sum_i\alphahat_i \kerneld(\cdot,X_i)$
where $\alphahat$ is the solution of~\eqref{eqn:krrDefn}.
While at first sight the differences with \krrs might seem superficial, 
we will see that 
the \emph{stronger} additive assumption will help us reduce the
excess risk for high dimensional regression.
Our theoretical results will be characterised directly via the optimisation
objective~\eqref{eqn:addObj}.

\subsection{The ESP Kernel}

While the above formulation reduces the number of optimisation parameters,
the kernel still has a combinatorial number of terms which can be expensive to
compute.
% computing the kernel is still problematic due to the large number of terms in the
% summation. 
While this is true for arbitrary choices for $\kernelj$'s, under some restrictions
we can efficiently compute $\kerneld$.
For this, we use the same trick used by~\citet{shawe2004kernel}
and \citet{duvenaud11additivegps}.
% However, under some restrictions on our kernel $\kerneld$, 
% we can efficiently compute it. 
First consider a set of base kernels acting on each dimension
$\kerneld_1,\kerneld_2,\dots,\dots,\kerneld_D$. 
Define $\kernelj$ to be the
product kernel of all kernels acting on each coordinate -- 
$\kernelj(\xj,{\xj}') = \kerneld_{i_1}(x_{i_1}, x'_{i_1}) 
\kerneld_{i_2}(x_{i_2}, x'_{i_2}) \cdots \kerneld_{i_d}(x_{i_d}, x'_{i_d})$.
Then, the additive kernel $\kerneld(x,x')$ becomes the $d$\superscript{th} 
elementary symmetric
polynomial (ESP) of the $D$ variables $\kerneld_1(x_1, x'_1), \dots, 
\kerneld_D(x_D, x'_D)$.
Concretely,
\begin{equation}
\kerneld(x,x') = %\sum_{j=1}^{M_d} \kernelj(\xj,{\xj}').
 \sum_{1\leq i_1<i_2<\dots<i_d\leq D} 
  \left( \prod_{\ell =1}^d \kerneld_{i_\ell} (x_{i_\ell}, x'_{i_\ell}) \right) .
\label{eqn:espkernel}
\end{equation} 
We refer to~\eqref{eqn:espkernel} as the ESP kernel. 
Using the Girard-Newton identities~\cite{macdonald95polynomials} for ESPs, we
can compute this summation efficiently.
For the $D$ variables $s_1^D = s_1,\dots,s_D$ and $1\leq m \leq D$, 
define the $m$\superscript{th} power sum $p_m$ and the $m$\superscript{th} elementary
symmetric polynomial $e_m$:
\ifthenelse{\boolean{istwocolumn}}{
\begin{align*}
p_m(s_1^D) &= \sum_{i=1}^D s_i^m \;, \\ %\hspace{0.2in} \textrm{where,}\\
e_m(s_1^D) &= \sum_{1\leq i_1<i_2<\dots<i_m\leq D}  
  s_{i_1} \times s_{i_2} \times \cdots \times s_{i_m}.
%   \prod_{\ell =1}^k s_{i_\ell} .
\end{align*}
}{
\[
p_m(s_1^D) = \sum_{i=1}^D s_i^m, \hspace{0.5in}
e_m(s_1^D) = \sum_{1\leq i_1<i_2<\dots<i_m\leq D}  
  s_{i_1} \times s_{i_2} \times \cdots \times s_{i_m}.
%   \prod_{\ell =1}^k s_{i_\ell} .
\]
}
In addition define $e_0(s_1^n) = 1$.
Then, the Girard-Newton formulae state,
\[
e_m(s_1^D) = \frac{1}{m} \sum_{i=1}^m (-1)^{i-1} e_{m-i}(s_1^D) p_i(s_1^D).
\]
Starting with $m=1$ and proceeding up to $m=d$,
$e_d$ can be computed iteratively  
in just $\bigO(Dd^2)$ time.
By treating $s_i = \kerneld_i$, the kernel matrix can be computed in
$\bigO(n^2 d^2 D)$ time.
% \citet{duvenaud11additivegps} use a similar technique in Gaussian processes.
While the ESP trick restricts the class of kernels we can use in \salsa,
it applies for important kernel choices.
For example, if each $\kernelj$ is a Gaussian kernel, then it 
is an ESP kernel if we set the bandwidths appropriately.

% We refer the resulting method \salsa.
In what follows, we refer to a kernel such as $\kerneld$~\eqref{eqn:espkernel}
which permits only  $d$ orders of interaction
as a $d$\superscript{th} order kernel. 
A kernel which permits interactions of all $D$ variables is of
$D$\superscript{th} order.
Note that unlike in MKL, here we do not wish to \emph{learn} the kernel.
We use additive kernels to explicitly reduce the complexity of the
function class over which we optimise for $\funchat$.
Next, we present our theoretical results.

\subsection{Theoretical Analysis}

We first consider the setting when $\regfuncj$ is in $\Hcalkj$ over which we optimise
for $\funchatj$.
Theorem~\ref{thm:addkrrthm} generally bounds the excess risk of 
$\funchat$~\eqref{eqn:addObj} in terms of RKHS parameters. Then, we
specialise it to specific RKHSs in Theorem~\ref{thm:kernelRates} 
and show that in many cases, 
the dependence on $D$ reduces from exponential to polynomial
for additive $\regfunc$.
We begin with some assumptions. \\[\thmparaspacing]

\begin{assumption}
$\regfunc$ has a decomposition
$\regfunc(x) = \sum_{j=1}^\Md \guncj(\xj)$ where each $\guncj \in \Hcalkj$
% and $\|guncj\|_\Hcalkj$ is bounded
.
\label{asm:decompAssumption}
\end{assumption}
\vspace{\thmparaspacing}

We point out that the decomposition $\{\guncj\}$ need not be unique. To
enforce definiteness (by abusing notation) 
we define $\regfuncj \in \Hcalkj$, $j =1,\dots,\Md$ 
to be the set of functions 
which minimise $\sum_{j} \|\guncj\|^2_\Hcalkj$. 
Denote the minimum value by 
$\|\regfuncvec\|^2_\Hcalaug$. 
% $\|\regfuncvec\|^2_\Hcalaug = \sum_j \|\regfuncj\|^2_\Hcalkj$. 
We denote it by a norm for
reasons made clear in our proofs.

Let $\kernelj$ have an eigenexpansion $\kernelj(\xj,{\xj}') = \sum_{\ell=1}^\infty 
\mulj \philj(\xj)\philj({\xj}')$ in $\Ltwo(\PXj)$.
Here, $\{(\philj)_{\ell=1}^\infty\}$ is an orthonormal basis for $\Ltwo(\PP_{\Xj})$
and $\{(\mulj)_{\ell=1}^\infty\}$ are its eigenvalues. $\PP_{\Xj}$ is the
marginal distribution of the coordinates $\Xj$.
We also need the following regularity condition on the tail
behaviour of the basis functions $\{\philj\}$ for all $\kernelj$. 
Similar assumptions are made in
\cite{zhang13dividekrr} and are satisfied for a large range of kernels including those
in Theorem~\ref{thm:kernelRates}.
\\[\thmparaspacing]

\begin{assumption}
For some $q\geq 2$, $\;\exists\;\rho < \infty$ such that
for all $j=1,\dots,M_d$ and $\ell \in \NN$, 
$\EE[\philj(X)^{2q}] \leq \rho^{2q}$.
\label{asm:basisTail}
\end{assumption}

We also define the following,
\begin{align*}
% \|\regfuncvec\|^2_\Hcalaug = \sum_{j=1}^\Md \|\regfuncj\|^2_\Hcalkj, \hspace{0.2in}
\gammaj(\lambda) = \sum_{\ell=1}^\infty \frac{1}{1 +  \lambda/\mulj}, \hspace{0.1in}
\gammakd(\lambda) = \sum_{j=1}^\Md \gammaj(\lambda).
\numberthis \label{eqn:termDefns}
\end{align*}
% The first quantity is the sum of all RKHS norms which we have denoted as a norm for
% reasons made clear in our proofs. 
The first term is known as the effective data
dimensionality of $\kernelj$ \cite{zhang13dividekrr,zhang05learning} and captures the
statistical difficulty of estimating a function in $\Hcalkj$. 
$\gammakd$ is the sum of the $\gammaj$'s.
% Under certain regularity conditions on the basis functions
% $\{{(\philj)_{\ell=1}^\infty}\vphantom{\}}_{j=1}^\Md\}$ 
% (See Assumption~\ref{asm:basisTail} in
% Appendix~\ref{sec:appMainThm})
Our first theorem below bounds the excess risk of $\funchat$ in terms
$\fnormsq$ and $\gammakd$.
\\[\thmparaspacing]

\begin{theorem}
Let Assumptions~\ref{asm:decompAssumption}
and~\ref{asm:basisTail} hold.
% Suppose $\regfunc$ satisfies the additive assumption of~\eqref{eqn:addAssumption}
% where each $\regfuncj \in \Hcalkj$. 
and $Y$ have bounded conditional variance:
$\EE[(Y-\regfunc(X))^2|X] \leq \sigma^2$. Then the solution $\funchat$
of~\eqref{eqn:addObj} satisfies,
\[
\EE[\Rcal(\funchat)] - \Rcal(\regfunc) \leq 
\Md\left( 20 \lambda \|\regfuncvec\|^2_\Hcalaug + \frac{12 \sigma^2 \gammakd(\lambda)}{n}
+ \chi(\kerneld) \right).
\]
\label{thm:addkrrthm}
\end{theorem}
\vspace{\thmparaspacing}
\vspace{\thmparaspacing}
Here $\chi(\kerneld)$ are kernel dependent low order terms and are 
given in~\eqref{eqn:chi} in Appendix~\ref{sec:appMainThm}.
Our proof technique generalises the analysis of~\citet{zhang13dividekrr}
for \krrs to the additive case.
We use ideas from~\citet{aronszajn50rkhs} to handle sum RKHSs.
We consider a space $\Hcalaug$ containing the tuple of functions 
$\funcii{j}\in \Hcalkj$
and use first order optimality conditions of~\eqref{eqn:addObj} in $\Hcalaug$.
%  By an analysis that follows \cite{zhang13dividekrr,zhang05learning} 
% we arrive at the above result. 
The proof is given in Appendix~\ref{sec:appMainThm}.

The term $\gammakd(\lambda)$, which typically has exponential dependence
on $d$, arises through the variance calculation. Therefore, by using small $d$
we may reduce the variance of our estimate.
However, this will also mean that we are only considering a smaller function class 
and
hence suffer large bias \emph{if} $\regfunc$ is not additive.
% In the above theorem, the optimal rate of convergence is obtained by balancing 
% $\lambda$ against $\gammakd(\lambda)/n$.
In naive \krr, using a $D$\superscript{th} order kernel (equivalent to
setting $\Md = M_D = 1$) the excess risk depends exponentially  in $D$.
In contrast, for an additive $d$\superscript{th} order kernel, $\gammakd(\lambda)$ 
has
polynomial dependence on $D$ \emph{if} $\regfunc$ is additive. 
We make this concrete via the following theorem.
\\[\thmparaspacing]

\begin{theorem}
Assume the same conditions as Theorem~\ref{thm:addkrrthm}.
Then, suppressing $\log(n)$ terms,
\begin{itemize}[leftmargin=*]
\item if each $\kernelj$ has eigendecay $\mulj \in \bigO(\ell^{-2s/d})$,
then by choosing $\lambda \asymp n^{\frac{-2s}{2s+d}}$, we have
$\EE[\Rcal(\funchat)] - \Rcal(\regfunc) \in \bigO(D^{2d} n^{\frac{-2s}{2s+d}})$,
\item if each $\kernelj$ has eigendecay 
$\mulj \in\bigO(\tilde{\pi}^d\exp(-\alpha\ell^2))$ for some constants
$\tilde{\pi},\alpha$, then by choosing $\lambda \asymp 1/n$, we have
$\EE[\Rcal(\funchat)] - \Rcal(\regfunc) \in \bigO(\frac{D^{2d} \tilde{\pi}^{d} }{n})$.
\end{itemize}
\label{thm:kernelRates}
\end{theorem}
% 
% \begin{theorem}
% Assume the same conditions as Theorem~\ref{thm:addkrrthm}. Let the kernels
% $\kernelj$ be obtained via the $d$-products of the base kernels
% $\kerneld_1,\dots,\kerneld_D$~\eqref{eqn:espkernel}.
% Then, suppressing $\log(n)$ terms,
% \begin{itemize}[leftmargin=*]
% \item if each base kernel $\kerneld_i$ has eigendecay 
% compact support and is $m$ times
% differentiable, then by choosing $\lambda \asymp n^{-1 + \frac{d}{m+2d}}$, we have
% $\EE[\Rcal(\funchat)] - \Rcal(\regfunc) \in \bigO(D^{2d} n^{-1+\frac{d}{m+2d}})$,
% \item if each base kernel $\kerneld_i$ is a Gaussian $\kerneld_i(x_i,x'_i) =
%   A \exp(-(x_i-x'_i)^2/2h_i^2)$, then by choosing $\lambda \asymp 1/n$, we have
% $\EE[\Rcal(\funchat)] - \Rcal(\regfunc) \in \bigO(\frac{D^{2d} (2\pi)^{d/2} }{n})$.
% \end{itemize}
% \label{thm:kernelRates}
% \end{theorem}

% \vskip{\thmparaspacing}
% For both kernels we use bounds on the eigenvalues given in~\citet{kuhn87eigvals}
% and~\citet{williamson01generalization}. 
We bound $\gammakd$ via bounds for $\gammaj$ and
use it to derive the optimal rates for the problem.
The proof is in Appendix~\ref{sec:appRates}.
% The $D^{2d}$ term appears due to the ${D \choose d}$
% kernels in the summation.

It is instructive to compare the rates for the cases above when we use a
$D$\superscript{th} order kernel $\kernel$ in \krrs to estimate a non-additive function.
The first eigendecay is obtained if each $\kernelj$ is a \matern kernel.
Then $\funcj$ belongs to the Sobolev class
of smoothness $s$~\cite{tsybakov08nonparametric,berlinet04RKHS}. 
By following a similar analysis, we can show that if $\kernel$ is in a Sobolev class,
then the excess risk of \krrs is $\bigO(n^{\frac{-2s}{2s+D}})$
which is significantly slower than ours. 
In our setting, the rates are only exponential in $d$ but we have an additional 
${D^{2d}}$ term as we need to estimate several such functions.
An example of the second eigendecay is the Gaussian kernel with
$\tilde{\pi}=\sqrt{2\pi}$~\cite{williamson01generalization}.
In the nonadditve case,
the excess risk is in the Gaussian RKHS is
$\bigO\big(\frac{(2\pi)^{D/2} }{n}\big)$ which is slower than \addkrrs 
whose dependence on $D$ is just polynomial.
$D, d$ do not appear in the exponent of $n$ because the
Gaussian RKHS contains very smooth functions.
\krrs is slower since we are optimising over the very large class of non-additive
functions and consequently it is a difficult statistical problem.
The faster rates for \addkrrs should not be surprising since the class of
additive functions is smaller.
The advantage of \addkrrs is its ability to recover the function at a faster rate
when $\regfunc$ is additive.
Finally we note that by taking each base kernel $\kerneld_i$ in the ESP kernel to 
be a 1D Gaussian, each $\kernelj$ is a Gaussian. However, at this point
it is not clear to us if it is possible to recover a $s$-smooth Sobolev class via the
tensor product of $s$-smooth one dimensional kernels.

Finally, we analyse \salsas under more agnostic assumptions. We will neither assume
that $\regfunc$ is additive nor that it lies in any RKHS. 
First, define the functions $\funclambdaj$, $j=1,\dots,M$
which minimise the population objective.
\ifthenelse{\boolean{istwocolumn}}{
\begingroup
\allowdisplaybreaks
\begin{align*}
&\{\funclambdaj\}_{j=1}^{\Md} = 
\argmin_{\funcj \in \Hcalkj, j = 1,\dots, M} 
\;\;\lambda \, \sum_{j=1}^\Md \|\funcj\|^2_{\Hcalkj} \;+ \\
&\hspace{1.1in}
  \EE\Bigg[\Big(Y - \sum_{j=1}^{\Md} \funcj (\Xj) \Big)^2\Bigg].
\numberthis \label{eqn:flambdaDefn}
\end{align*}
\endgroup
}{
\begin{align*}
\{\funclambdaj\}_{j=1}^{\Md} = 
\argmin_{\funcj \in \Hcalkj, j = 1,\dots, M} 
\;\;\lambda \, \sum_{j=1}^\Md \|\funcj\|^2_{\Hcalkj} \;+\;
% \hspace{1.1in}
  \EE\Bigg[\Big(Y - \sum_{j=1}^{\Md} \funcj (\Xj) \Big)^2\Bigg].
\numberthis \label{eqn:flambdaDefn}
\end{align*}
}
Let $\funclambda = \sum_j\funclambdaj$,  $\Rlambdaj = \|\funclambdaj\|_\Hcalkj$ 
and $\Rlambda^2 =\sum_j{\Rlambdaj}^2$. 
To bound the excess risk in the agnostic setting we also define
the class,
\ifthenelse{\boolean{istwocolumn}}{
\begin{align*}
\Fcallambda = \big\{&f:\Xcal\rightarrow\RR; 
  \;\;f(x)=\sum_j\funcj(\xj), 
  \numberthis \label{eqn:RcalDefn} \\[-0.05in]
  &\; \forall j, \;\funcj \in \Hcalkj, \|\funcj\|_\Hcalkj \leq
\Rlambdaj\big\}.
\end{align*}
}{
\begin{align*}
\Fcallambda = \big\{f:\Xcal\rightarrow\RR; 
  \;\;f(x)=\sum_j\funcj(\xj), 
  \; \forall j, \;\funcj \in \Hcalkj, \|\funcj\|_\Hcalkj \leq
\Rlambdaj\big\}.
  \numberthis \label{eqn:RcalDefn} 
\end{align*}
}

\begin{theorem}
Let $\regfunc$ be an arbitrary %(possibly non-additive) 
measurable function and $Y$ have bounded
fourth moment $\EE[Y^4]\leq\nu^4$. Further each $\kernelj$ satisfies
Assumption~\ref{asm:basisTail}. Then $\forall\;\eta>0$,
\ifthenelse{\boolean{istwocolumn}}{
\begingroup
\allowdisplaybreaks
\begin{align*}
&\EE[\Rcal(\funchat)] - \Rcal(\regfunc) \leq (1+\eta)\approxErr \;+\;
  (1 + 1/\eta)\estimErr, \\
% &\textrm{where, }\;\;
&{\rm where, }\;
\approxErr = \inf_{f\in\Fcallambda} \|f-\regfunc\|_2^2, \hspace{0.1in}
\estimErr \in \bigO\Big(\frac{M_d \gammakd(\lambda)}{n} \Big).
\end{align*}
\endgroup
}{
\begin{align*}
\Rcal(\funchat) - \Rcal(\regfunc) 
\;\;\leq \;\; (1+\eta)\underbrace{\inf_{f\in\Fcallambda} \|f-\regfunc\|_2^2}_{\approxErr} 
  \;\;+\;\;
 (1 + 1/\eta)\underbrace{\bigO\Big(\frac{M_d \gammakd(\lambda)}{n}\Big)}_{\estimErr} 
\end{align*}
}
\label{thm:agnosticThm}
\end{theorem}
\vspace{\thmparaspacing}
\vspace{\thmparaspacing}
\vspace{\thmparaspacing}
The proof, given in Appendix~\ref{sec:appAgnostic}, also follows the template
in~\citet{zhang13dividekrr}.
Loosely, we may interpret $\approxErr$ and $\estimErr$ as the approximation and
estimation errors\footnote{Loosely (and not strictly) since $\funchat$ need not be 
in $\Fcallambda$.}.
We may use Theorem~\ref{thm:agnosticThm} to understand the trade-offs in
% approximating non-additive $\regfunc$ via additive $\funchat$.
approximaing a non-additive function via an additive model.
We provide an intuitive ``not-very-rigorous" explanation.
$\Fcallambda$ is typically increasing with $d$ since higher order additive 
functions contain lower order functions.
Hence, $\approxErr$ is decreasing with $d$ as the infimum is taken over a larger
set.
On the other hand, $\estimErr$ is increasing with $d$.
% due to both $\Md$ and $\gammakd(\lambda)$.
% since both $\Md$  and $\gammakd(\lambda)$ are exponential in $d$.
With more data $\estimErr$ decreases due to the $1/n$ term.
Hence, we can afford to use larger $d$ to reduce $\approxErr$ and balance
with $\estimErr$. This results in an overall reduction in the excess risk.
% $\Rlambda^2$ is also generally increasing since $\Fcallambda$ is increasing.
% Hence, $\estimErr$ is increasing with $d$.
%  when $\Md\gammakd(\lambda)\Rlambda^2$ is increasing 
% If $d_1 < d_2$, then $\Fcalclass_{d_1,\lambda}\subset\Fcalclass_{d_1,\lambda}$
% as any model which permits $d_2$ interactions also permits $d_1$
% interactions. Therefore, $\approxErr$ is decreasing with $d$.
% However, $M_{d_1}\gammakd(\lambda) < M_{d_2}\gammakd(\lambda)$
% since both $\Md$ and $\gammakd(\lambda)$ increase exponentially in $d$.
% Therefore, $\estimErr$ is increasing with $d$.
% The exponential to polynomial improvement is reflected in the $\Md\gammakd(\lambda)$
% term in $\estimErr$.
% Note that $R_{\lambda_1} > R_{\lambda_2}$ and $
% \Fcal_{\lambda_1}\supset\Fcal_{\lambda_2}$ when $\lambda_1<\lambda_2$.
% Hence $\approxErr$ increases with $\lambda$ while $\estimErr$ decreases.

The actual analysis would be more complicated since $\Fcallambda$ is a bounded
class depending intricately on $\lambda$. 
It also depends on the kernels $\kernelj$, which differ with $d$.
To make the above intuition concrete and more interpretable, it is necessary to have a 
good handle on $\approxErr$. 
% We may bound $\approxErr$ when $\regfunc$ is in a nonparametric function class
% such as the H\"older and Sobolev classes~\cite{tsybakov08nonparametric}.
However, if we are to overcome the exponential dependence in dimension,
usual nonparametric assumptions such as H\"olderian/ Sobolev conditions alone 
will not suffice. Current lower bounds
suggest that the exponential dependence is 
unavoidable~\cite{gyorfi02distributionfree,tsybakov08nonparametric}. 
Additional assumptions will be necessary to demonstrate faster convergence.
Once we control $\approxErr$,
% we may optimise the bound over $\eta,\lambda$ to obtain the optimal rates.
the optimal rates can be obtained by optimising the bound over $\eta, \lambda$.
We wish to pursue this in future work.

%!TEX root = v2hastie.tex

\subsection{Practical Considerations}

\textbf{Choice of Kernels: }
The development of our algorithm and our analysis
assume that the $\kerneld_i$'s are known. This is hardly the case in reality  and
they have to be chosen properly for good empirical performance.
Cross validation is not feasible here as there are too many
hyper-parameters. 
In our experiments we set each $\kerneld_i$ to be a Gaussian kernel 
$\kerneld_i(x_i,x'_i) = \sigma_Y \exp(-(x_i-x'_i)^2/2h^2_i)$ with 
bandwidth $h_i = c \sigma_{i} n^{-1/5}$.
Here $\sigma_{i}$ is the standard deviation of the $i$\superscript{th}
covariate and $\sigma_Y$ is the standard deviation of $Y$.
The choice of bandwidth was inspired by several other kernel methods
which use bandwidths on the order
$\sigma_{i} n^{-1/5}$ \cite{tsybakov08nonparametric,ravikumar09spam}.
The constant $c$ was hand tuned -- we found that 
performance was robust to choices between $5$ and $60$.
In our experiments we use $c=20$.
% In our experiments we set each $\kerneld_i$ to be a Gaussian kernel 
% $\kerneld_i(x_i,x'_i) = A \exp(-(x_i-x'_i)^2/2h^2_i)$ with 
% bandwidth $h_i = c \sigma_{i} n^{-1/5}$ and scale $A = (\sigma_Y/M_d)^{1/d}$.
% Here $\sigma_{i}$ is the standard deviation of the $X$'s on the $i$\superscript{th}
% dimension and $\sigma_Y$ is the standard deviation of $Y$.
% The choice of bandwidth was inspired by several other kernel methods
% which use bandwidths on the order
% $\sigma_{i} n^{-1/5}$ \cite{tsybakov08nonparametric,ravikumar09spam}.
% The constant $c$ was hand tuned -- we found that the
% performance of \addkrrs was robust to choices between $5$ and $60$.
% In our experiments we use $c=20$. The parameter $A$ was set so that 
% $\kerneld$ satisfies $\kerneld(x,x) = \sigma_Y$ to capture the
% scale of variation in the regressand.
$c$ was chosen by experimenting on a collection of synthetic
datasets and then used in all our experiments. Both synthetic and real datasets
used in experiments are independent of the data used to tune $c$.

\textbf{Choice of $d,\lambda$: }
If the additive order of $\regfunc$ is known and we have sufficient data then we
can use that for $d$ in~\eqref{eqn:espkernel}. However, this
is usually not the case in practice.
Further, even in non-additive settings, we may wish to use an additive model
to improve the variance of our estimate.
In these instances, our approach to choose $d$ uses cross validation.
For a given $d$ we solve~\eqref{eqn:krrDefn} for different $\lambda$ and 
pick the best one via cross validation. 
To choose the optimal $d$ we cross validate on $d$. 
In our experiments we observed that the cross validation error had 
bi-monotone like behaviour with a unique 
local optimum on $d$. Since the optimal $d$ was typically small we search 
by starting at $d=1$ and keep increasing until the error begins to increase
again.
If $d$ could be large and linear search becomes too expensive, a binary search like 
procedure on $\{1,\dots,D\}$ can be used.

We conclude this section with a couple of remarks. 
First, we could have considered an alternative additive model which 
sums all interactions up to
$d$\superscript{th} order instead of just the $d$\superscript{th} order.
The excess risk of this model differs from
% by subdominant terms and/or constant factors from 
Theorems~\ref{thm:addkrrthm},~\ref{thm:kernelRates} and~\ref{thm:agnosticThm}
only in subdominant terms and/or constant factors.
The kernel can be computed efficiently using the same trick by
summing all polynomials up to $d$.
In our experiments we found that both our original
model~\eqref{eqn:addAssumption} and summing over all interactions 
performed equally well.
For simplicity, results are presented only for the former.

Secondly, as is the
case with most kernel methods, \salsas requires $\bigO(n^2)$ space to store
the kernel matrix and $\bigO(n^3)$ effort to invert it.
% Secondly, as is the
% case with most kernel methods, \salsas requires $\bigO(n^2)$ space and
% $\bigO(n^3)$ effort to store and invert the kernel matrix.
Some recent advances in scalable kernel methods such as random
features, divide and conquer techniques, stochastic gradients 
etc.~\cite{rahimi07random,rahimi08kitchen,le13fastfood,zhang13dividekrr,dai14scalable} 
can be explored to scale \salsas with $n$.
% One could explore 
% some recent advances in kernel methods which use random
% features~\cite{rahimi07random,rahimi08kitchen,le13fastfood} or 
% divide-and-conquer techniques~\cite{zhang13dividekrr} to scale \salsas with $n$.
However, this is beyond the scope of this paper and is left to future work.
For this reason, we also limit our experiments to moderate dataset sizes.
The goal of this paper is primarily to introduce additive models of higher order,
address the combinatorial cost in such models and theoretically demonstrate 
the improvements in the excess risk.

% \vspace{-0.06in}
\section{Experiments}
\label{sec:experiments}
% \vspace{-0.06in}

% \textbf{Alternatives for Comparison: } 
We compare \addkrrs to the following.
\textbf{Nonparametric models:} Kernel Ridge Regression (\krr),
$k$-Nearest Neighbors (\knn), 
Nadaraya Watson (\nw),
Locally Linear/ Quadratic interpolation (\locallin, \localquad),
$\epsilon$-Support Vector Regression (\sveps),
$\nu$-Support Vector Regression (\svnu),
Gaussian Process Regression (\gp),
Regression Trees (\rt),
Gradient Boosted Regression Trees (\gbrt) \cite{friedman00gbrt},
RBF Interpolation (\rbfi),
M5' Model Trees (\mfp) \citep{wang97m5prime}
and 
Shepard Interpolation (\si).
\textbf{Nonparametric additive models:}
Back-fitting with cubic splines (\backf) \citep{hastie90gam},
Multivariate Adaptive Regression Splines (\mars) \citep{friedman91mars},
Component Selection and Smoothing (\cosso) \citep{lin2006cosso},
Sparse Additive Models (\spam) \citep{ravikumar09spam}
and 
Additive Gaussian Processes (\addgp) \citep{duvenaud11additivegps}.
\textbf{Parametric models:}
Ridge Regression (\lr),
Least Absolute Shrinkage and Selection (\lasso) \cite{tibshirani94lasso} and
Least Angle Regression (\lar) \cite{efron04lar}.
We used software from 
\cite{chang11libsvm,jakabsons15regression,rasmussen06gps,lin2006cosso,
hara13humanPose} or 
from Matlab. In some cases we used our own implementation.
\insertFigToy

\vspace{-0.1in}
\subsection{Synthetic Experiments}
% \vspace{-0.05in}

We begin with a series of synthetic examples.
We compare \salsas to some non-additive methods to convey intuition about our additive
model.
% We compare only to subset of the above methods to avoid clutter in the figures.
First we create a synthetic low order function of order $d=3$ in $D=15$ 
dimensions.
We do so by creating a $d$ dimensional function $f_d$ and add that function over all
${D\choose d}$ combinations of coordinates. We compare \addkrrs using order $3$ and
compare against others. The results are given in
Figure~\ref{fig:knownorder}.  
This setting is tailored to the assumptions of our method and, not surprisingly, it
outperforms all alternatives.

Next we demonstrate the bias variance trade-offs in using additive 
approximations 
on non-additive functions. We created a $15$ dimensional (non-additive) function and
fitted a \addkrrs model with $d = 1, 2, 4, 8, 15$ for difference
choices of $n$. The results are given in Figure~\ref{fig:comporder}.
The interesting observation here is that for small samples sizes small $d$ 
performs best. However, as we increase the sample size we can also increase the
capacity of the model by accommodating higher orders of interaction. In this regime,
large $d$ produces the best results. 
This illustrates our previous point that the
order of the additive model gives us another way to control the bias and variance
in a regression task.
We posit that when $n$ is extremely large,
$d=15$ will eventually  beat all other models.  
Finally, we construct synthetic functions in $D=20$
to $50$ dimensions and compare  against other methods in Figures~\ref{fig:cvone}
to~\ref{fig:cvfour}. Here, we chose $d$ via cross validation.
Our method outperforms or is competitive with other methods.

\vspace{-0.05in}
\subsection{Real Datasets}
% \vspace{-0.1in}

Finally we compare \addkrrs against the other methods listed above on 16 datasets. The
datasets were taken from the UCI repository, Bristol Multilevel Modeling and 
the following sources: \citep{tegmark06lrgs,wehbe14brain,just10neurosemantic,
guillame14utility,tu14pancreatic,paschou07pca}.
Table~\ref{tb:realData} gives the average squared error on a test set.
% The dimensionality and number of training points is indicated next to the dataset.
% The best method(s) for each dataset are in bold. The second to fifth methods 
% are underlined. 
For the Speech dataset we have indicated the training time (including
cross validation for selecting hyper-parameters) for each method.
For \addkrrs we have also indicated the order $d$ chosen by cross validation.
See the caption under the table for more details.
% The datasets with a * are actually lower dimensional datasets from the UCI repository. 
% But we
% artificially increase the dimensionality by inserting random values for the remaining
% coordinates. Even though, this doesn't change the function value it makes the
% regression problem harder.

\addkrrs performs best (or is very close to the best) in 5 of the datasets.
Moreover it falls within the top $5$ in all but two datasets, coming sixth in both
instances. 
% It is by far the most consistent of the 22 methods. 
% In the table we have also indicated the order $d$ of the \addkrrs model 
% chosen via cross validation. 
Observe that
in many cases $d$ chosen by \salsas is much smaller than $D$, but \emph{importantly} 
also larger than $1$.
This observation (along with Fig~\ref{fig:comporder})
corroborates a key theme of this paper: while it is true that additive models 
improve the variance in high dimensional regression, it is often insufficient to
confine ourselves to just first order models.

In Appendix~\ref{sec:appExpDetails} we have given the specifics on the datasets such
as preprocessing, the predictors, features etc.
We have also discussed some details on the alternatives used.

% \textbf{Some experimental details: }
% \gps is the Bayesian interpretation of \krr. However, the results are
% different in Table~\ref{tb:realData}. We believe this is due to differences in 
% hyper-parameter tuning. For \gp, the 
% GPML package~\cite{rasmussen06gps} optimises the
% \gps marginal likelihood via L-BFGS. In contrast, our \krrs implementation
% minimises the least squares cross validation error via grid search.
% Some \addgps results are missing since it was very slow compared to 
% other methods.
% On the Blog dataset, \addkrrs took less than $35$s to train and all other
% methods were completed in under 22 minutes. In contrast \addgps was not done training
% even after several hours. Even on the relatively small speech dataset 
% \addgps took about $80$ minutes.
% Among the others, \backf, \mars, and \spams were the more expensive methods requiring
% several minutes on datasets with large $D$ and $n$ 
% whereas other methods took under 2-3 minutes.
% We also experimented with locally cubic and quartic interpolation but exclude
% them from the table since \locallin, \localquads generally performed better.
% Appendix~\ref{sec:appExpDetails} has more details on the synthetic functions
% and test sets.

% Insert the big table here so that its the last page of the paper
% ================================================================
\insertTableReal
% ================================================================
%!TEX root = v2hastie.tex

% \vspace{-0.1in}
\section{Conclusion}
% \vspace{-0.1in}
\label{sec:conclusion}

% \textbf{Summary: }
\addkrrs finds additive approximations to the regression function in
high dimensions.
It has less bias than first order models and less variance than non-additive methods.
% Unlike first order methods, we do not 
% considerably compromise on the statistical integrity of the model.
Algorithmically, it requires plugging in an additive kernel
to \krr. In computing the kernel,
we use the Girard-Newton formulae to efficiently sum over
a combinatorial number of terms.
Our theorems show that the excess risk
depends only polynomially on $D$ 
% for  \addkrrs has only polynomial dependence on $D$
when $\regfunc$ is additive, significantly better than the usual
exponential dependence of nonparametric methods,
% results on nonparametric regression which have exponential dependence, 
albeit under stronger assumptions.
% We also analyse the agnostic setting.
Our analysis of the agnostic setting provides intuitions on the tradeoffs invovled 
with changing $d$. %%%% added for long version
We demonstrate the efficacy of \salsas via a comprehensive
empirical evaluation.
Going forward, we wish to 
use techniques from scalable kernel methods to handle large datasets.

% \textbf{Future work: }
% \vspace{-0.1in}
Theorems~\ref{thm:addkrrthm},\ref{thm:kernelRates} show polynomial
dependence on $D$ when $\regfunc$ is additive. 
However, these theorems are unsatisfying since in practice regression functions
need not be additive.
% However, these results have limited applicability
% when $\regfunc$ is not additive.
We believe our method did well even on non-additive settings 
since we could control model capacity via $d$.
In this light, we pose the following open problem:
identify suitable assumptions to beat existing lower bounds and prove
faster convergence of additive models whose additive order $d$ increases with 
sample size $n$.
Our Theorem~\ref{thm:agnosticThm} might be useful in this endeavour.
%  is an exciting open problem arising out of our work.

% \vspace{0.2in}
\subsection*{Acknowledgements}
We thank Calvin McCarter, Ryan Tibshirani and Larry Wasserman for the insightful
discussions and feedback on the paper. We also thank Madalina Fiterau for providing
us with datasets. This work was partly funded by DOE grant DESC0011114.

\vspace{0.2in}
\subsection*{Acknowledgements}
We thank Calvin McCarter, Ryan Tibshirani and Larry Wasserman for the insightful
discussions and feedback on the paper. We also thank Madalina Fiterau for providing
us with datasets. This work was partly funded by DOE grant DESC0011114.

% \newpage
\vspace{0.1in}
\appendix
\section*{\LARGE Appendix}

% \section{Convergence of \addkrr when in an RKHS}
\section{Proof of Theorem~\ref{thm:addkrrthm}: Convergence of \salsa}
% \section{Proof of Theorem~\ref{thm:addkrrthm}}
\label{sec:appMainThm}

% This section presents the proof of Theorem~\ref{thm:addkrrthm}. 
Our analysis here is a brute force generalisation of the analysis 
in~\citet{zhang13dividekrr}.
We handle the additive case using ideas from~\citet{aronszajn50rkhs}.
As such we will try and stick to the same notation.
Some intermediate technical results can be obtained directly 
from~\citet{zhang13dividekrr} but
we repeat them (or provide an outline) here for the sake of completeness.

In addition to the definitions presented in the main text, we will also need the
following quantities,
\begin{align*}
\betajt = \sum_{\ell = t+1}^\infty \mulj, \hspace{0.2in}
% \betat = \sum_{j=1}^\Md \betajt \hspace{0.2in}
\Psij = \sum_{\ell=1}^\infty \muj_\ell, \hspace{0.2in}
\bntq = \max\left( \sqrt{\max(q,\log t)} \, ,\, \frac{\max(q,\log t)}{n^{1/2 - 1/q}}
\right).
\end{align*}
Here $\Psij$ is the trace of $\kernelj$.
$\betajt$ depends on some $t \in \NN$ which we will pick later.
Also  define $\betat = \sum_j\betajt$ and $\Psi = \sum_j \Psij$.

Note that the excess risk can be decomposed into bias and
variance terms, $\Rcal(\funchat) - \Rcal(\regfunc) = 
\EE[\|\funchat - \regfunc\|_2^2] =
\|\regfunc - \EE\funchat\|_2^2 + \EE[\|\funchat - \EE\funchat\|_2^2]$.
In Sections~\ref{sec:appBias} and~\ref{sec:appVariance} respectively, we will prove
the following bounds which will yield in Theorem~\ref{thm:addkrrthm}:
\begin{align*}
\|\regfunc - \EE\funchat\|_2^2 &\leq M_d\Bigg( 8 \lambda \fnormsq
  + \frac{8 \Md^{3/2} \rho^4 \fnormsq }{\lambda} \Psi \betat
      + \fnormsq \sum_{j=1}^\Md \muj_{t+1} \numberthis \label{eqn:bias} %\\
%   &\hspace{0.4in}
  + \Big(\frac{C\Md b(n,t,q)\rho^2\gammakd(\lambda)}{\sqrt{n}}\Big)^q \ftwonormsq
  \Bigg), \\
\EE[\|\funchat - \EE\funchat\|_2^2]
&\leq \Md\Bigg( 12\lambda \fnormsq +
\frac{12\sigma^2\gammakd(\lambda)}{n}  + \numberthis \label{eqn:variance} \\
&\hspace{0.4in}
\left(\frac{2\sigma^2}{\lambda} + 4\fnormsq\right) 
\bigg( \sum_{j=1}^\Md \muj_{t+1} + \frac{12\Md\rho^4}{\lambda} \Psi \betat
  + \Big(\frac{C\Md b(n,t,q)\rho^2\gammakd(\lambda)}{\sqrt{n}}\Big)^q\bigg)
  \ftwonormsq \Bigg).
\end{align*}
Accordingly, this gives the following expression for $\chi(\kerneld)$,
\begin{align*}
\chi(\kerneld) &=  \inf_{t}\Bigg[ \frac{8 \Md^{3/2} \rho^4 \fnormsq }{\lambda} \Psi \betat 
 + \left(\frac{2\sigma^2}{\lambda} + 4\fnormsq + 1\right) 
\bigg(\frac{C\Md b(n,t,q)\rho^2\gammakd(\lambda)}{\sqrt{n}}\bigg)^q\ftwonormsq \Bigg) +
\\ &\hspace{0.4in}
\left(\frac{2\sigma^2}{\lambda} + 4\fnormsq\right) 
\bigg( \sum_{j=1}^\Md \muj_{t+1} + \frac{12\Md\rho^4}{\lambda} \Psi \betat \bigg) 
+ \fnormsq \sum_{j=1}^\Md \muj_{t+1}
\Bigg].
\numberthis \label{eqn:chi}
\end{align*}
Note that the second term in $\chi(\kerneld)$ is usually low
order  for large enough $q$
due to the $n^{-q/2}$ term. Therefore if in our setting
$\betajt$ and $\muj_{t+1}$ are small enough, $\chi(\kerneld)$ is low order. We show
this for the two kernel choices of Theorem~\ref{thm:kernelRates} 
in Appendix~\ref{sec:appRates}.

First, we review some well known results on RKHS's which we will use in our analysis.
Let $\kernel$ be a PSD kernel and $\Hcalk$ be its RKHS.
Then $\kernel$ acts as the representer of evaluation -- i.e. for any $f \in \Hcalk$, 
$\rinner{f}{\kernel(\cdot, x)}_\Hcalk = f(x)$.
Denote the RKHS norm $\|f\|_\Hcalk = \sqrt{\rinner{f}{f}_\Hcalk}$ and the 
$\Ltwo$ norm $\|f\|_2 = \sqrt{\int f^2}$. 

Let the kernel $\kernel$ have an eigenexpansion $\kernel(x,x') = \sum_{\ell=1}^\infty
\mu_\ell \phi_\ell(x) \phi_\ell(x')$. Denote the basis coefficients of $f$ in
$\{\phi_\ell\}$ via $\{\theta_\ell\}$. 
That is, $\theta_\ell = \int f \cdot \phi_\ell \, \ud\PP$ and $f = \sum_{\ell=1}^\infty
\theta_\ell \phi_\ell$.
The following results are well known~\cite{scholkopf01kernels,steinwart08svms},
\[
\rinner{\phi_\ell}{\phi_\ell} = 1/\mu_\ell, \hspace{0.2in}
\|f\|^2_{2} = \sum_{\ell=1}^\infty \theta_\ell^2, \hspace{0.2in}
\|f\|^2_\Hcalk = \sum_{\ell=1}^\infty \frac{\theta_\ell^2}{\mu_\ell}.
\]

Before we proceed, we make the following remark on the minimiser
of~\eqref{eqn:addObj}.
\\[\thmparaspacing]
% \newpage

\begin{remark}
The solution of~\eqref{eqn:addObj} takes the form 
$\funchat(\cdot) = \sumin\alpha_i\kerneld(\cdot, X_i)$ where $\kerneld$ is the sum
kernel~\eqref{eqn:sumkernel}.
\vspace{\thmparaspacing}
\label{rmk:sumRKHS}
\end{remark}
\begin{proof}
The key observation is that we only need to consider $n$ (and not $n\Md$) parameters
 even though we are optimising over $\Md$ RKHSs.
The reasoning uses a powerful result from~\citet{aronszajn50rkhs}.
Consider the class of functions $\sumrkhs' = \{f = \sumj\funcj;\funcj\in\Hcalkj\}$.
In~\eqref{eqn:addObj} we are minimising over $\sumrkhs'$.
Any $f\in\sumrkhs'$ need \emph{not} have a unique additive decomposition.
Consider $\sumrkhs\subset\sumrkhs'$ which only contains the minimisers in the
expression below.
% The following is a norm in $\sumrkhs$, 
\[
\|f\|^2_\sumrkhs = \inf_{\guncj\in\Hcalkj; f =\sum\guncj} 
\sumjM \|\guncj\|^2_\Hcalkj
\]
\citet{aronszajn50rkhs} showed that $\Hcal$ is an RKHS with the sum kernel
$\kerneld = \sumj \kernelj$ and its RKHS norm is $\|\cdot\|_\sumrkhs$.
% The infimum is necessary since any $f\in\Hcal$ need not have a unique decomposition. 
Clearly, the minimiser of~\eqref{eqn:addObj} lies in $\sumrkhs$. For any
$g'\in\sumrkhs'$, we can pick a corresponding $g\in\sumrkhs$ with the same sum of
squared errors (as $g=g'$) but lower complexity penalty (as $g$ minimises
the sum of norms for any $g'=g$).
Therefore, we may optimise~\eqref{eqn:addObj} just over $\sumrkhs$ and not
$\sumrkhs'$. An application of Mercer's theorem concludes the proof.
\end{proof}

\vspace{0.1in}
\subsection{Set up}
\label{sec:thmSetup}

We first define the following function class of the product of all RKHS's,
$
\Hcalaug = 
\Hcalkii{1}\times\Hcalkii{2}\times\dots\times\Hcalkii{\Md} =
\left\{ \funcvec = (\funcone, \dots, \funcMd) \big|
\funcj \in \Hcalkj \; \forall j \right\}
$
and equip it with the inner product
$
\inner{\funcvec_1}{\funcvec_2} = 
% {\inner{\funcone_1, \funcone_2}}_{\Hcalkii{1}}
\rinner{\funcone_1}{\funcone_2}_{\Hcalkii{1}} + \dots + 
\rinner{\funcMd_1}{\funcMd_2}_{\Hcalkii{M_d}} .
$
Here, $\funcj_1$ are the elements of $\funcvec_1$ and
$\rinner{\cdot}{\cdot}_{\Hcalkj}$ is the RKHS inner product of $\Hcalkj$.
Therefore the norm is $\|\funcvec\|_\Hcalaug^2 = \sum_{j=1}^\Md
\|\funcii{j}\|^2_\Hcalkj$.
% It is clear that $\Hcalaug$ is a RKHS with kernel 
% % Let
% $\kernelaug(x,x') = \big(\kernelii{1}(\xii{1}, {\xii{1}}'), \dots ,
% \kernelii{\Md}(\xii{\Md}, {\xii{\Md}}')\big)$.
Denote $\xijx = \kernelj(x, \cdot)$ and $\xi_x(\cdot) = \kernelaug(\cdot, x)$.
Observe that for an additive function $\func = \sum_j \funcj(x)$,
\[
\func(x) = \sum_j \funcj(x) = \sum_j \rinner{\funcj}{\xijx}_\Hcalkj 
= \rinner{\funcvec}{\xix}.
\]
Recall that the solution to~\eqref{eqn:addObj} is denoted by $\funchat$ and
the individual functions of the solution are given by $\funchatj$.
We will also use $\regfuncvec$ and $\funchatvec$ to denote the representations of
$\regfunc$ and $\funchat$ in $\Hcalaug$, i.e.,
$\regfuncvec = (\regfuncii{1}, \dots, \regfuncii{\Md})$ and 
$\funchatvec = (\funchatii{1}, \dots, \funchatii{\Md})$.
Note that $\|\regfuncvec\|^2_\Hcalaug$ is precisely the bound used in
Theorem~\ref{thm:addkrrthm}.
We will also denote $\Deltaj = \funchatj - \regfuncj \in \Hcalkj$, 
$\Deltavec = (\Deltaii{1},\dots,\Deltaii{\Md}) \in \Hcalaug$,
and $\Delta = \sum_j \Deltaj = \funchat - \regfunc$.

For brevity, from now on we will write $\kernelj(x,x')$ instead of
$\kernelj(\xj,{\xj}')$. Further, since $d$ is fixed in this analysis
we will write $M$ for $\Md$.

\subsection{Bias (Proof of Bound~\eqref{eqn:bias})}
\label{sec:appBias}

Note that we need to bound $\|\EE[\Delta]\|_2$ which by Jensen's inequality is less
than $\EE[\|\EE[\Delta|\Xn]\|_2]$. 
Since, $\|\EE[\Delta|\Xn]\|_2^2
\leq M \sum_{j=1}^M \|\EE[\Deltaj|\Xn]\|_2^2$, we will focus on bounding 
$ \sum_{j=1}^M \|\EE[\Deltaj|\Xn]\|_2^2$.

We can write the optimisation objective~\eqref{eqn:addObj} as follows,
\begin{equation}
\funchatvec = \argmin_{\funcvec \in \Hcalaug}
  \frac{1}{n} \sum_{i=1}^n \left( \rinner{\funcvec}{\xi_{X_i}} - Y_i\right)^2
  + \lambda \|\funcvec\|^2_\Hcalaug
\label{eqn:Fobj}
\end{equation}
Since this is Fr\'{e}chet differentiable in $\Hcalaug$ in
the metric induced by the inner product defined above, the first order optimality 
conditions for $\funchatj$ give us,
\[
\frac{1}{n} \sum_{i=1}^n \left( \rinner{\xi_{X_i}}{\funchatvec - \regfuncvec}
  -\epsilon_i \right) \xij_{X_i} \;+\; 2\lambda \funchatj = \zero.
\]
Here, we have taken $Y_i = \regfunc(X_i) + \epsilon_i$ where $\EE[\epsilon_i|X_i] =
0$. 
Doing this for all $\funchatj$  we have,
\begin{equation}
\frac{1}{n}\sum_{i=1}^n \xi_{X_i} \left(\rinner{\xi_{X_i}}{\Deltavec} - \epsilon_i
\right) + \lambda \funchatvec = \zero
\label{eqn:optCond}
\end{equation}
Taking expectations conditioned on $\Xn$ and rearranging we get,
\begin{align*}
(\Sigmahat + \lambda I)\EE[\Deltavec| \Xn] = - \lambda \regfuncvec,
\numberthis
\label{eqn:optCondtwo}
\end{align*}
where $\Sigmahat = \frac{1}{n}\sum_i \xi_{X_i} \otimes \xi_{X_i}$ is the empirical
covariance.
Since $\Sigmahat \succeq \zero$,
\begin{align}
\forall j', \;\;\;
\|\EE[\Deltaii{j'}|\Xn]\|^2_{\Hcalkii{j'}} \leq
\sum_{j=1}^M 
\|\EE[\Deltaj|\Xn]\|_\Hcalkj^2  = \|\EE[\Deltavec|\Xn]\|^2_\Hcalaug \leq 
\|\regfuncvec\|^2_\Hcalaug
\label{eqn:condRKHSBound}
\end{align}

Let $\EE[\Deltaj|\Xn] = \sum_{\ell=1}^\infty \deltalj \philj$ where $\philj$ are the
eigenfunctions in the expansion of $\kernelj$.
Denote $\deltajdown = (\deltaj_1,\dots,\deltaj_t)$ and
$\deltajup = (\deltaj_{t+1},\deltaj_{t+2},\dots )$. 
We will set $t$ later.
Since $\|\EE[\Deltaj|\Xn]\|_2^2 = \|\deltajdown\|_2^2 + \|\deltajup\|_2^2$
we will bound the two terms. The latter term is straightforward,
\begin{align}
\|\deltajup\|_2^2 \leq \muj_{t+1} \sum_{\ell = t+1}^\infty \frac{{\deltalj}^2}{\mulj}
\leq \muj_{t+1} \|\EE[\Deltaj|\Xn]\|^2_\Hcalkj \leq \muj_{t+1}\fnormsq
\label{eqn:biasLatterTerm}
\end{align}
To control $\|\deltajdown\|$, let $\regfuncj = \sum_\ell \thetalj \philj$.
Also, define the following:
$\thetajdown = (\thetaj_1,\dots,\thetaj_t)$,
$\Phij \in\RR^{n\times t}$, $\Phij_{i\ell} = \philj(X_i)$,
$\Philj = (\philj(X_1),\dots,\philj(X_n)) \in \RR^n$,
$\Mj = \diag(\muj_1,\dots,\muj_t) \in \RR_+^{t\times t}$ and
$\vj\in \RR^n$ where $\vj_i = \sum_{\ell>t} \deltalj \philj(X_i) =
\EE[\Deltajup(X_i)|\Xn]$. 

Further define,
$\Phi = [\Phiii{1} \dots \Phiii{M}] \in \RR^{n\times tM}$,
$\Mcal = \diag(\Mii{1},\dots,\Mii{M}) \in \RR^{tM \times tM}$, 
$v_i = \sum_j \vj$, $\deltadown = [\deltaiidown{1}; \dots; \deltaiidown{M}] \in
\RR^{tM}$ and $\thetadown = [\thetaiidown{1}; \dots; \thetaiidown{M}] \in \RR^{tM}$.

Now compute the $\Hcalaug$-inner product between $(\zero,\dots,\philj,\dots,\zero)$ with 
equation~\eqref{eqn:optCondtwo} to obtain,
\begin{align*}
\frac{1}{n}\sum_{i=1}^n \rinner{\philj}{\xij_{X_i}}_\Hcalkj
\rinner{\xi_{X_i}}{\EE[\Deltavec|\Xn]} + \lambda
\rinner{\philj}{\EE[\Deltaj|\Xn]}_\Hcalkj
&= -\lambda \rinner{\philj}{\regfuncj}_\Hcalkj \\
\frac{1}{n} \sum_{i=1}^n \philj(X_i) 
\sum_{j=1}^M \left(\sum_{\ell'\leq t} \philpj(X_i)\deltalpj 
  + \sum_{\ell'>t} \philpj(X_i)\deltalpj\right) + \lambda \frac{\deltalj}{\mulj}
  &= -\lambda\frac{\thetalj}{\mulj}
\end{align*}
After repeating this for all $j$ and for all $\ell=1,\dots,t$, 
and arranging the terms appropriately this reduces to
\[
\left( \frac{1}{n}\Phi^\top \Phi + \lambda \Mcal^{-1}\right) \deltadown
= - \lambda \Mcal^{-1} \thetadown - \frac{1}{n} \Phi^\top v
\]
By writing $Q = (I + \lambda \Mcal^{-1})^{1/2}$, we can rewrite the above expression
as
\[
\left(I + Q^{-1}\left(\frac{1}{n}\Phi^\top \Phi - I\right)Q^{-1} \right)
Q\deltadown = -\lambda Q^{-1}\Mcal^{-1}\thetadown - \frac{1}{n}Q^{-1}\Phi^\top v.
\]

We will need the following technical lemmas. The proofs are given at the end of this
section. These results correspond to Lemma 5 in~\citet{zhang13dividekrr}.
\\[\thmparaspacing]

% We now use the following technical result from~\citet{zhang13dividekrr} (Lemma 5). 
% We skip the proof by noting that its derivation almost exactly is the same as the
% proof of the lemma in \cite{zhang13dividekrr} with only a couple of
% differences. The first is that we need to
% perform the steps over $\Hcalaug$ instead of an ordinary Hilbert space.
% The second is that additional $M$ terms appear when we use Jensen's
% inequality to bound the sum over $M$ terms.
% \\[\thmparaspacing]

% \begin{lemma}[Modified from \citet{zhang13dividekrr}]
% The following bounds are true:
% \begin{align*}
% \|\lambda Q^{-1} M^{-1} \thetadown\|_2^2 = 
% \lambda\sum_{j=1}^M \|{Q^{(j)}}^{-1}{\Mj}^{-1}\thetadown\|_2^2 
%   &\leq \lambda \fnormsq \\
% \EE\left[\|\frac{1}{n}Q^{-1}\Phi^\top v\|_2^2\right]
% \leq \frac{M\rho^4\fnormsq\Psi \betat}{\lambda}
% \end{align*}
% Further, Define the event $\Ecal = \{\|Q^{-1}(\frac{1}{n}\Phi^\top \Phi - I)
% Q^{-1}\|_{op} \leq 1/2\}$. Then, there exists $C$ such that 
% \[
% \PP(\Ecal^c) \leq \left(\max\left( \sqrt{\max(q,\log t)} \;, \; \frac{\max(q,\log
% t)}{n^{1/2-1/q}} \right) \times \frac{MC\rho^2\gammakd(\lambda)}{\sqrt{n}} \right)^q
% \]
% \label{lem:technicallemma}
% \end{lemma}

\begin{lemma}
$ \|\lambda Q^{-1} \Mcal^{-1} \thetadown\|_2^2 
  \leq \lambda \fnormsq$.
\\[\thmparaspacing]
\label{lem:technicalOne}
\end{lemma}

\begin{lemma}
$ \EE\left[\|\frac{1}{n}Q^{-1}\Phi^\top v\|_2^2\right]
\leq \frac{1}{\lambda}M^{3/2}\rho^4\fnormsq\Psi \betat$.
\\[\thmparaspacing]
\label{lem:technicalTwo}
\end{lemma}

\begin{lemma}
Define the event $\Ecal = \{\|Q^{-1}(\frac{1}{n}\Phi^\top \Phi - I)
Q^{-1}\|_{op} \leq 1/2\}$. Then, there exists a constant $C$ s.t.
\[
\PP(\Ecal^c) \leq \left(\max\left( \sqrt{\max(q,\log t)} \;, \; \frac{\max(q,\log
t)}{n^{1/2-1/q}} \right) \times \frac{MC\rho^2\gammakd(\lambda)}{\sqrt{n}} \right)^q.
\]
\\[\thmparaspacing]
\label{lem:technicalThree}
\end{lemma}

When $\Ecal$ holds, by Lemma~\ref{lem:technicalThree} and noting that $Q \succeq I$,
\begin{align*}
\|\deltadown\|^2_2 \leq \|Q\deltadown\|_2^2
  &= \Big\|
  \left(I + Q^{-1}\left(\frac{1}{n}\Phi^\top \Phi - I\right)Q^{-1} \right)^{-1}
  \left(-\lambda Q^{-1}\Mcal^{-1}\thetadown - \frac{1}{n}Q^{-1}\Phi^\top v\right)
  \Big\|^2 \\
&\leq 4\|\lambda Q^{-1}\Mcal^{-1}\thetadown
+ \frac{1}{n}Q^{-1}\Phi^\top v\|_2^2.
  \leq 8\|\lambda Q^{-1}\Mcal^{-1}\thetadown\|_2^2
+ 8\|\frac{1}{n}Q^{-1}\Phi^\top v\|_2^2 
\end{align*}
% Here we have applied Lemma~\ref{lem:technicalThree}. 
Now using Lemmas~\ref{lem:technicalOne} and~\ref{lem:technicalTwo},
\[
\EE[\|\deltadown\|_2^2|\Ecal] \leq 8\left( \lambda\fnormsq + 
 \frac{M^{3/2}\rho^4\fnormsq\Psi \betat}{\lambda} \right)
\]
Since
$ \EE[\|\deltadown\|_2^2]
= \PP(\Ecal) \EE[\|\deltadown\|_2^2 | \Ecal] + 
\PP(\Ecal^c)\EE[\|\deltadown\|_2^2|\Ecal^c]
$
and by using the fact that $\|\deltadown\|^2 \leq \|\EE[\Delta|\Xn]\|_2^2 \leq
\ftwonormsq$, we have
\begin{align*}
\EE[\|\deltadown\|_2^2] &\leq 8\lambda\fnormsq + 
 \frac{8 M\rho^4\fnormsq\Psi \betat}{\lambda} \; +
 \\ &\hspace{0.3in} 
  \left(\max\left( \sqrt{\max(q,\log t)} \;, \; \frac{\max(q,\log
t)}{n^{1/2-1/q}} \right) \times \frac{MC\rho^2\gammakd(\lambda)}{\sqrt{n}} \right)^q
  \ftwonormsq
\end{align*}
% Since $\EE[\|\deltadown\|_2^2|\Ecal^c]$

Finally using~\eqref{eqn:biasLatterTerm} and by noting that 
\begin{align*}
\|\EE[\Delta|\Xn]\|_2^2 \;\;\leq M \sum_{j=1}^M \|\EE[\Deltaj|\Xn]\|_2^2
\;\;= M\big( \|\deltadown\|_2^2 + \sum_j \|\deltajup\|_2^2 \big)
\;\;\leq M\big( \|\deltadown\|_2^2 + \fnormsq \sum_j \muj_{t+1}  \big)
\end{align*}
and then taking expectation over $\Xn$,
we obtain the bound for the bias in~\eqref{eqn:bias}.

\vspace{0.2in}
\subsection*{Proofs of Technical Lemmas}

\subsubsection{Proof of Lemma~\ref{lem:technicalOne}}
Lemma~\ref{lem:technicalOne} is straightforward.
\begin{align*}
\|Q^{-1}\Mcal^{-1} \thetadown\|_2^2 &= 
    \sum_{j=1}^M \|{\Qj}^{-1} {\Mj}^{-1} \thetajdown\|_2^2 
%   = \sum_{j=1}^M {\thetajdown}^\top {\Mj}^{-1}(I + \lambda {\Mj}^{-1})^{-1}
%       {\Mj}^{-1} \thetajdown \\
  = \sum_{j=1}^M {\thetajdown}^\top ({\Mj}^{2} + \lambda {\Mj})^{-1}
      \thetajdown \\
  &\leq \sum_{j=1}^M {\thetajdown}^\top (\lambda {\Mj})^{-1}
      \thetajdown
  = \frac{1}{\lambda}\sum_{j=1}^M \sum_{\ell =1}^t \frac{{\thetalj}^2}{\mulj}
  \leq \frac{1}{\lambda}\|\regfuncvec\|^2_\Hcalaug
\end{align*}

\subsubsection{Proof of Lemma~\ref{lem:technicalTwo}}

We first decompose the LHS as follows,
\begin{align}
\left\|\frac{1}{n}Q^{-1}\Phi^\top v\right\|_2^2
  = \left\|(M + \lambda I)^{-1/2}\left(\frac{1}{n}M^{1/2}\Phi^\top v\right) 
          \right\|_2^2
  \leq \frac{1}{\lambda}\left\| \frac{1}{n}M^{1/2}\Phi^\top v \right\|_2^2
  \numberthis \label{eqn:biasQPVdecomp}
\end{align}
The last step follows by noting that $\|(M+\lambda I)^{-1/2}\|^2_{op}
= \max_{j,\ell} 1/(\mulj + \lambda) \leq 1/\lambda$.
Further,
\begin{align*}
\EE\left[\|M^{1/2}\Phi^\top v\|_2^2\right] =
  \sum_{j=1}^M \sum_{\ell = 1}^t \mulj \EE[({\Philj}^\top v)^2]
  \leq \sum_{j=1}^M \sum_{\ell = 1}^t \mulj \EE[\|\Philj\|_2^2 \|v\|_2^2]
  \numberthis \label{eqn:biasMPVbound}
%   \leq \sum_{j=1}^M \sum_{\ell = 1}^t \mulj 
\end{align*}
Note that the term inside the summation in the RHS can be bounded by,
$\sqrt{\EE[\|\Philj\|^4_2]\EE [\|v\|^4_2]}$. We bound the first expectation via,
\[
\EE\left[\|\Philj\|^4\right]
  = \EE\left[\bigg(\sum_{i=1}^n\philj(X_i)^2\bigg)^2\right]
  \leq \EE\left[n\sum_{i=1}^n\philj(X_i)^4\right] \leq n^2\rho^4
\]
where the last step follows from Assumption~\ref{asm:basisTail}. For the second
expectation we first bound $\|v\|^4$,
\[
\|v\|_2^4 = \left(\sum_{i=1}^n \bigg(\sum_{j=1}^M \vj_i\bigg)^2\right)^2
  \leq \left(M\sum_{i=1}^n \sum_{j=1}^M {\vj_i}^2\right)^2
  \leq M^3n\sum_{i=1}^n \sum_{j=1}^M {\vj_i}^4
\]
Now by the Cauchy Schwarz inequality,
\[{\vj_i}^2 = \bigg(\sumlgt \deltalj\philj(X_i)\bigg)^2
\leq \bigg(\sumlgt\frac{{\deltalj}^2}{\mulj}\bigg)
\bigg(\sumlgt\mulj\philj(X_i)^2\bigg).
% \leq \fnormsq \bigg(\sumlgt\mulj\philj(X_i)^2\bigg).
\]
Therefore,
\begin{align*}
\EE\left[\|v\|^4\right] &\leq M^3n \sumin\sumjM
  \EE\left[ \|\EE[\Deltaj|\Xn]\|_\Hcalkj^4 \bigg(\sumlgt\mulj\philj(X_i)^2\bigg)^2 \right]
  \\
  &\leq M^3n\fnorm^4 \sumjM\sumin \sum_{\ell,\ell'>t} 
  \EE[\mulj\mu^{(j)}_{\ell'} \philj(X_i)^2\phi^{(j)}_{\ell'}(X_i)^2] \\
  &\leq M^3n\rho^4\fnorm^4 \sumjM\sumin \left(\sum_{\ell>t} \mulj\right)^2
  \leq M^3n^2\rho^4\fnorm^4\sumjM{\betajt}^2
\end{align*}
Here, in the first step we have used the definition of
$\|\EE[\Deltaj|\Xn]\|_\Hcalkj$,
in the second step, equation~\eqref{eqn:condRKHSBound}, in the third step
assumption~\ref{asm:basisTail} and Cauchy Schwarz, and in the last step, the
definition of $\betat$.
Plugging this back into~\eqref{eqn:biasMPVbound}, we get
\[
\EE\left[\|M^{1/2}\Phi^\top v\|^2\right] \leq
  M^{3/2}n^2\rho^4\fnormsq\sqrt{\sumjM{\betajt}^2}\sumjM\sumlt\mulj
  \leq 
  M^{3/2}n^2\rho^4\fnormsq\Psi\betat
\]
This bound, along with equation~\eqref{eqn:biasQPVdecomp} gives us the desired result.

\subsubsection{Proof of Lemma~\ref{lem:technicalThree}}

Define $\piij = \{\philj(x_i)\}_{\ell=1}^t \in \RR^t$,
$\pii = [\piiii{1};\dots;\piiii{M}] \in \RR^{tM}$ and the matrices 
$\Ai = Q^{-1}(\pii\pii^\top - I)Q^{-1} \in \RR^{tm\times tM}$. Note that $\Ai =
\Ai^\top$ and
\[
\EE[A_i] = Q^{-1}(\EE[\pii\pii^\top] - I)Q^{-1} =  \zero.
\]
Then, if $\epsilon_i, i = 1,\dots,n$ are \iid Rademacher random variables, by a
symmetrization argument we have,
\begin{align}
\EE\left[\Big\|Q^{-1}\left(\frac{1}{n}\Phi^\top\Phi - I\right)Q^{-1} \Big\|_{op}^k\right]
= \EE\left[\Big\| \frac{1}{n}\sum_{i=1}^n A_i  \Big\|_{op}^k \right]
\leq 2^k \EE\left[\Big\|\frac{1}{n}\sum_{i=1}^n\epsilon_i A_i \Big\|_{op}^k\right]
\label{eqn:rademacher}
\end{align}
The above term can be bounded by the following expression. 
\begin{align*}
&2^q\left( \sqrt{e\max(q,\log(t))} \frac{\rho^2\sqrt{M}}{\sqrt{n}}
\sqrt{\sum_{\ell=1}^M \gammaj(\lambda)^2}
+ 4e\max(q,\log(t))\rho^2 \left(\frac{M}{n}\right)^{1-1/q} 
 \left(\sum_{\ell=1}^M \gammaj(\lambda)^q\right)^{1/q}
\right)^q \\
&\leq \left(\frac{C}{2}\right)^q \max\left(\sqrt{M(\max(q,\log t))}, 
  \frac{M^{1-1/q}\max(q,\log t)}{n^{1/2-1/q}}
  \right)^q 
\left(\frac{\rho^2 \gammakd(\lambda)}{\sqrt{n}}\right)^q
\end{align*}
The proof mimics Lemma 6 in~\cite{zhang13dividekrr} by performing essentially the
same steps over $\Hcalaug$ instead of the usual Hilbert space. In many of the steps,
$M$ terms appear (instead of the one term for \krr) which is accounted for via
Jensen's inequality. 

Finally, by Markov's inequality,
\begin{align*}
\PP(\Ecal^c) &\leq 2^k 
\EE\left[\Big\|Q^{-1}\left(\frac{1}{n}\Phi^\top\Phi - I\right)Q^{-1} 
  \Big\|_{op}^q\right] \\
&\leq C^q \max\left(\sqrt{M(\max(q,\log t))}, 
  \frac{M^{1-1/q}\max(q,\log t)}{n^{1/2-1/q}}
  \right)^q 
\left(\frac{\rho^2 \gammakd(\lambda)}{\sqrt{n}}\right)^q
\end{align*}

\subsection{Variance (Proof of Bound~\eqref{eqn:variance})}
\label{sec:appVariance}

Once again, we follow \citet{zhang13dividekrr}. 
The tricks we use to
generalise it to the additive case (i.e. over $\Hcalaug$) are the same as that for
the bias. %Hence, we will only provide an outline here.
Note that since $\EE[\|\funchat - \EE\funchat\|_2^2] 
\leq \EE[\|\funchat - g\|_2^2]$ for all $g$, it is sufficient to
bound $\EE[\|\funchat - \regfunc\|_2^2] = \EE[\|\Delta\|_2^2]$.

First note that,
\begin{align*}
\lambda \EE[\|\funchatvec\|^2_\Hcalaug|\Xn] \;\;\leq
  \EE\left[ \frac{1}{n}\sum_{i=1}^n \left( \funchat(X_i) - Y_i \right)^2 + 
      \lambda\|\funchatvec\|^2_\Hcalaug \bigg| \Xn \right] 
  \;\;\leq \frac{1}{n} \sum_{i=1}^n \EE[\epsilon_i^2|\Xn] + \lambda \fnormsq
  \leq \sigma^2 + \lambda \fnormsq
\end{align*}
The second step follows by the fact that $\funchatvec$ is the minimiser
of~\eqref{eqn:Fobj}. Then, for all $j$,
\begin{equation}
\EE[\|\Deltaj\|_\Hcalkj^2|\Xn] \leq
\EE[\|\Deltavec\|^2_\Hcalaug|\Xn] 
\leq 2\fnormsq + 2\EE[\|\funchatvec\|_2^2|\Xn] 
\leq \frac{2\sigma^2}{\lambda} + 4\fnormsq
\label{eqn:Deltabound}
\end{equation}

Let $\Deltaj = \sum_{\ell =1}^\infty \deltalj \philj$. Note that the definition of
$\deltalj$ is different here. Define
$\deltajdown, \deltajup, \Deltajdown, \Deltajup, \deltadown$ analogous to the
definitions in Section~\ref{sec:appBias}.
Then similar to before we have,
\[
\EE[\|\deltajup\|_2^2] \leq \muj_{t+1} \EE[\|\Deltajup\|^2_\Hcalkj] 
\leq \muj_{t+1}\left(\frac{2\sigma^2}{\lambda} + 4\fnormsq \right)
\]
We may use this to obtain a bound on $\EE[\|\Deltaup\|^2]$. To obtain a bound on
$\EE[\|\Deltadown\|^2]$,
take the $\Hcalaug$ inner product of $(\zero,\dots,\philj,\dots,\zero)$ with
the first order optimality condition~\eqref{eqn:optCond} and following essentially
the same procedure to the bias we get,
\[
\left(\frac{1}{n}\Phi^\top \Phi + \lambda \Mcal^{-1} \right) \deltadown
= - \lambda \Mcal^{-1} \thetadown - \frac{1}{n}\Phi^\top v +
\frac{1}{n}\Phi^\top \epsilon
\]
where $\Phi,\Mcal,\thetadown$ are the same as in the bias calculation.
$\vj\in \RR^n$ where $\vj_i = \sum_{\ell>t} \deltalj \philj(X_i) =
\EE[\Deltajup(X_i)|\Xn]$ (recall that $\deltalj$ is different to the definition in the
bias) and
$\epsilon \in \RR^n$, $\epsilon_i = Y_i - \regfunc(X_i)$ is the vector of errors. 
Then we write,
\[
\left(I + Q^{-1}\left(\frac{1}{n}\Phi^\top \Phi - I\right) Q^{-1} \right)
Q\deltadown = -\lambda Q^{-1}\Mcal^{-1}\thetadown - \frac{1}{n}Q^{-1}\Phi^\top v
+ \frac{1}{n}Q^{-1}\Phi^\top \epsilon
\]
where $Q = (I+\lambda\Mcal^{-1})^{1/2}$ as before.
Following a similar argument to the bias, when the event $\Ecal$ holds,
\begin{align*}
\|\deltadown\|_2^2 \leq \|Q\deltadown\|_2^2 &\leq 4\|\lambda Q^{-1}\Mcal^{-1}\thetadown + 
\frac{1}{n}Q^{-1}\Phi^\top v + \frac{1}{n}Q^{-1}\Phi^\top \epsilon \|_2^2 \\
 &\leq 12\|\lambda Q^{-1}\Mcal^{-1}\thetadown\|^2 + 
12 \|\frac{1}{n}Q^{-1}\Phi^\top v \|^2 + 12 \|\frac{1}{n}Q^{-1}\Phi^\top \epsilon \|_2^2 
% &\leq 12 \lambda \fnormsq + \frac{12M\rho^2\Psi\betat (2\sigma^2/\lambda +
% 4\fnormsq)}{\lambda} +
% \frac{12\sigma^2 \gammakd(\lambda)}{n}
\numberthis \label{eqn:varBoundOne}
\end{align*}
By Lemma~\ref{lem:technicalOne}, the first term can be bounded via $12 \lambda
\fnormsq$. For the second and third terms we use the following two lemmas, the proofs
of which are given at the end of this subsection.
\\[\thmparaspacing]

\begin{lemma}
$ \EE\left[ \|\frac{1}{n}Q^{-1}\Phi^\top v \|_2^2\right] \leq 
\frac{1}{\lambda}M\rho^4\Psi\betat (2\sigma^2/\lambda + 4\fnormsq)
$.
\\[\thmparaspacing]
\label{lem:technicalFour}
\end{lemma}

\begin{lemma}
$
\EE\left[ \big\|\frac{1}{n}Q^{-1}\Phi^\top \epsilon\big\|_2^2 \right]
\leq  \frac{\sigma^2}{n} \gammakd(\lambda)
$
\\[\thmparaspacing]
\label{lem:technicalFive}
\end{lemma}

Note that
$\EE[\|\deltadown\|^2_2] \leq \PP(\Ecal) \EE[\|\deltadown\|_2^2|\Ecal] 
+ \EE[\indfone(\Ecal^c)\|\deltadown\|_2^2]$. 
The bound on the first term comes via equation~\eqref{eqn:varBoundOne} and
Lemmas~\ref{lem:technicalOne},~\ref{lem:technicalFour} and~\ref{lem:technicalFive}.
The second term can be bound via,
\begin{align*}
\EE[\indfone(\Ecal^c)\|\deltadown\|_2^2] &\leq
\EE[\indfone(\Ecal^c)\EE[\|\Deltavec\|_\Hcalaug^2|\Xn] \\
&\leq \left(\max\left( \sqrt{\max(q,\log t)} \;, \; \frac{\max(q,\log
t)}{n^{1/2-1/q}} \right) \times \frac{MC\rho^2\gammakd(\lambda)}{\sqrt{n}} \right)^q
\left( \frac{2\sigma^2}{\lambda} + 4\fnormsq \right)
\numberthis \label{eqn:varBoundTwo}
\end{align*}
Here, we have used equation~\eqref{eqn:Deltabound} and Lemma~\ref{lem:technicalThree}.
Finally, note that 
\begin{align*}
\EE[\|\Delta\|_2^2] &\leq M\sum_j \EE[\|\Deltaj\|_2^2]
= M\big( \EE\|\deltadown\|_2^2 + \sum_j \EE\|\deltajup\|_2^2 \big) \\
  &\leq M\bigg(\EE\|\deltadown\|_2^2 + \Big(\frac{2\sigma^2}{\lambda} + 4\fnormsq \Big) 
  \sum_j \muj_{t+1} \bigg)
\numberthis \label{eqn:varBoundThree}
\end{align*}
When we combine~\eqref{eqn:varBoundOne},~\eqref{eqn:varBoundTwo}
and~\eqref{eqn:varBoundThree} we get the bound in equation~\eqref{eqn:variance}.

\vspace{0.2in}
\subsection*{Proofs of Technical Lemmas}

\subsubsection{Proof of Lemma~\ref{lem:technicalFour}}
Note that following an argument similar to equation~\eqref{eqn:QPVdecomp} in
Lemma~\ref{lem:technicalTwo}, it is sufficient to bound $\EE\|M^{1/2}\Phi^\top v\|_2^2$.
We expand this as,
\begin{align*}
\EE\left[\|M^{1/2}\Phi^\top v\|_2^2\right]
  &= \sum_{j=1}^M \sum_{\ell=1}^t \mulj \EE[({\Philj}^\top v)^2] 
  \leq \sum_{j=1}^M \sum_{\ell=1}^t \mulj \EE[\|\Philj\|^2\|v\|^2]  
%   &= \sum_{j=1}^M \sum_{\ell=1}^t \mulj \EE\left[\|\Philj\|^2 \EE[\|v\|^2|\Xn]\right] 
\end{align*}
To bound this term, first note that 
\[
\|v\|^2 = \sum_{i=1}^n \bigg(\sum_{j=1}^M \vj_i \bigg)^2
\leq M \sum_{i=1}^n \sum_{j=1}^M {\vj_i}^2
\leq M \sum_{i=1}^n \sum_{j=1}^M 
%   \underbrace{\bigg(\sum_{\ell>t} \frac{{\deltalj}^2}{\mulj}\bigg)}_{\leq \|\Deltaj\|^2_\Hcalkj}
  \bigg(\sum_{\ell>t} \frac{{\deltalj}^2}{\mulj}\bigg)
  \bigg(\sum_{\ell>t} \mulj \philj(X_i)^2\bigg)
\]
Therefore, 
\begin{align*}
\EE\left[\|M^{1/2}\Phi^\top v\|^2\right]
&\leq \sumjM\sumlt \mulj M \sumin \sumjpM
  \EE\left[ \EE[\|\Deltaii{j'}\|^2_\Hcalkii{j'}|\Xn] \|\Philj\|^2 
    \sumlpgt\mu^{(j')}_{\ell'} \phi^{(j')}_{\ell'}(X_i)^2
  \right] \numberthis \label{eqn:MPvIntermediate} \\
&\leq M\left(\frac{2\sigma^2}{\lambda} + 4\fnormsq \right)
  \sumjM\sumlt\mulj \sumin\sumjpM\sumlpgt \mu^{(j')}_{\ell'}
  \EE\left[\|\Philj\|^2 \phi^{(j')}_{\ell'}(X_i)^2 \right]
\end{align*}
For all $i$, the inner expectation can be bounded using
assumption~\ref{asm:basisTail} and Jensen's inequality via,
\begin{align*}
\EE\left[\|\Philj\|^2\phi^{(j')}_{\ell'}(X_i)^2\right]
&\leq \sqrt{
  \EE\left[\|\Philj\|^4\right]\EE\left[\phi^{(j')}_{\ell'}(X_i)^4\right]}
  \leq \rho^2 \sqrt{\EE\left[\bigg(\sumin\philj(X_i)^2 \bigg)^2\right]} \\
& \leq \rho^2 \sqrt{ \EE\left[n \sumin\philj(X_i)^4 \right]} 
  \leq \rho^2 \sqrt{ n^2\rho^4} = n\rho^4.
\end{align*}
This yields,
\begin{align*}
\EE\left[\|M^{1/2}\Phi^\top v\|^2\right] &\leq 
 Mn^2\rho^4\left(\frac{2\sigma^2}{\lambda} + 4\fnormsq \right)
  \underbrace{\sumjM\sumlt\mulj}_{\leq\Psi} 
  \underbrace{\sumjpM\sumlpgt \mu^{(j')}_{\ell'}}_{=\betat}
% = Mn^2\rho^4\Psi\betat\left(\frac{2\sigma^2}{\lambda} + 4\fnormsq \right)
\end{align*}
Finally, we have
\begin{align}
\EE\left[\left\|\frac{1}{n}Q^{-1}\Phi^\top v\right\|_2^2\right]
  \leq \EE\left[\frac{1}{\lambda}\left\| \frac{1}{n}M^{1/2}\Phi^\top v \right\|_2^2
          \right]
  \numberthis \label{eqn:QPVdecomp}
  \leq \frac{1}{\lambda} M\rho^4\Psi\betat
  \left(\frac{2\sigma^2}{\lambda} + 4\fnormsq \right)
\end{align}

\subsubsection{Proof of Lemma~\ref{lem:technicalFive}}
We expand the LHS as follows to obtain the result.
\[
\EE\left[ \big\|\frac{1}{n}Q^{-1}\Phi^\top \epsilon\big\|^2 \right]
= \frac{1}{n^2}\sum_{j=1}^M \sum_{\ell=1}^t \sum_{i=1}^n
  \frac{1}{1 + \lambda/\mulj} \EE[{\philj(X_i)}^2 \epsilon_i^2 ]
\leq \frac{\sigma^2}{n} \sum_{j=1}^M \gammaj(\lambda)
= \frac{\sigma^2}{n} \gammakd(\lambda)
\]
The first step is just an expansion of the matrix. 
In the second step we have used 
$\EE[{\philj(X_i)}^2 \epsilon_i^2] =
 \EE[{\philj(X_i)}^2 \EE[\epsilon_i^2|X_i]]\leq \sigma^2$ since
$\EE[\philj(X)^2] = 1$. In the last two steps we
have used the definitions of $\gammaj(\lambda)$ and $\gammakd(\lambda)$.

\vspace{0.2in}
\section{Proof of Theorem~\ref{thm:kernelRates}: Rate of Convergence in Different
RKHSs}
% \section{Proof of Theorem~\ref{thm:kernelRates}}
\label{sec:appRates}

Our strategy will be to choose $\lambda$ so as to balance the 
dependence on $n$ in the first two terms in the RHS of the bound in
Theorem~\ref{thm:addkrrthm}.
% The term $\chi(\lambda)$ can be shown to be lower order even though we do not
% explicity discuss it here. We refer the interested reader to \cite{zhang13dividekrr}.

\begin{proof}[Proof of Theorem~\ref{thm:kernelRates}-1]
% \textbf{$\kerneld_i$ has compact support  and is $m$ times differentiable}:\\
\textbf{Polynomial Decay}:\\
% First consider a product kernel $\kernelt = \kerneld_1 \times \dots \kerneld_d$.
% Let the eigen expansion of $\kernelt$ be $\kernelt(x,x') = \sum_{\ell=1}^\infty 
% \mut_\ell \phi_\ell(x) \phi_\ell(x')$.
% If $\kerneld_i$ is a $m$ times differentiable and and has compact support,
% then the same could be said about $\kernelt$. For such kernels \citet{kuhn87eigvals}
% (Theorem 4) gives us the decay  $\mu_\ell \leq C \ell^{-m/d -1}$ for some constant $C$.
% 
% Since each $\kernelj$ in the sum of~\eqref{eqn:sumkernel} is $\kernelt$, 
The quantity
$\gammakd(\lambda)$ can be bounded via $M_d \sum_{\ell=1}^\infty 1/(1 +
\lambda/\mut_\ell) $. If we set $\lambda = n^{\frac{-2s}{2s+d}}$, then
\begingroup
\allowdisplaybreaks
\begin{align*}
\frac{\gammakd(\lambda)}{\Md} &=  \sum_{\ell=1}^\infty 
  \frac{1}{1 + n^{\frac{-2s}{2s+d}}/\mut_\ell }
  \;\;\leq\;\; n^{\frac{d}{2s+d}} + 
  \sum_{\ell > n^{\frac{d}{2s+d}}} \frac{1}{1 +
n^{\frac{2s}{2s+d}}\ell^{\frac{2s}{d}} } \\
  &\leq n^{\frac{d}{2s+d}} + n^{-\frac{2s}{2s+d}} \
  \sum_{\ell > n^{\frac{d}{2s+d}}} \frac{1}{n^{\frac{-2s}{2s+d}} + \ell^{\frac{2s}{d}}} \\
  &\leq n^{\frac{d}{2s+d}} + n^{\frac{-2s}{2s+d}} \
  \left( n^{\frac{d}{2s+d}} + 
      \int_{ n^{\frac{d}{2s+d}}}^\infty u^{-2s/d} \ud u \right)  
  \in \bigO(n^{\frac{d}{2s+d}}).
\end{align*}
\endgroup
Therefore, $\gammakd(\lambda)/n \in \bigO(\Md  n^{\frac{-2s}{2s+d}})$ giving the
correct dependence on $n$ as required. 
To show that $\chi(k)$ is negligible, set $t = n^{\frac{3d}{2s-d}}$.
Ignoring the $\textrm{poly}(D)$ terms,
both $\mut_{t+1}, \betat \in \bigO(n^{\frac{-6s}{2s-d}})$ and 
$\chi(\kerneld)$ is low order.
Therefore, by Thereom~\ref{thm:addkrrthm}
the excess risk is in $\bigO(M_d^2 n^{\frac{-2s}{2s+d}})$.
\end{proof}

\newcommand{\pit}{\tilde{\pi}}
\begin{proof}[Proof of Theorem~\ref{thm:kernelRates}-2]
\textbf{Exponential Decay}:\\
% \textbf{$\kerneld_i$ is Gaussian}:\\
% If each $\kerneld_i$ is a gaussian, then the product kernel 
% $\kernelt = \kerneld_1 \times \dots \kerneld_d$ is also a Gaussian. 
% From the work of~\citet{williamson01generalization} (Lemma 14 \& Remark 15) we know
% that the $d$ dimensional Gaussian kernel has eigendecay 
% $\mut_\ell \leq C \pit^d \exp(-\alpha \ell^2)$ where
% $\pit = \sqrt{2\pi}$ and $\alpha$ is a constant. 
By setting $\lambda = 1/n$ and following
a similar argument to above we have,
\begin{align*}
\frac{\gammakd(\lambda)}{\Md} &\leq
  \sqrt{\frac{\log n}{\alpha}} + \frac{1}{\lambda}
      \sum_{\ell>  \sqrt{\log n/\alpha} } \mut_\ell
  \;\;\leq  \;\;
  \sqrt{\frac{\log n}{\alpha}} + n\pit^d \sum_{\ell>  \sqrt{\log n/\alpha} } 
    \exp(-\alpha \ell^2)  \\
  &\leq 
  \sqrt{\frac{\log n}{\alpha}} +
  n\pit^d \left( \frac{1}{n} +   \int_{\sqrt{\log n/\alpha} }^\infty 
    \exp(-\alpha \ell^2)  \right) 
  =   \sqrt{\frac{\log n}{\alpha}}  +   \pit^d \left( 1 + \frac{\sqrt{\pi}}{2}(1 -
\Phi(\sqrt{\log n}) \right),
\end{align*}
where $\Phi$ is the Gaussian cdf. 
In the first step we have bounded the first $\sqrt{\frac{\log n}{\alpha}}$ terms by 
$1$ and then bounded the second term by a constant.
Note that the last term is $o(1)$.
Therefore ignoring $\log n$ terms, 
$\gammakd(\lambda) \in \bigO( M_d \pit^d)$ which gives excess risk
$\bigO(M_d^2 \pit^d/n)$.
$\chi(\kerneld)$ can be shown to be low order by choosing $t=n^2$ which results in
$\mut_{t+1}, \betat \in \bigO(n^{-4})$.
\end{proof}

\vspace{0.2in}
\section{Proof of Theorem~\ref{thm:agnosticThm}: Analysis in the Agnostic Setting}
\label{sec:appAgnostic}

As before, we generalise the analysis by~\citet{zhang13dividekrr} to the tuple
RKHS $\Hcalaug$.
We begin by making the following crucial observation about the population minimiser
\eqref{eqn:flambdaDefn}
$\funclambda = \sumjM\funclambdaj$,
\begin{equation}
\funclambda = \argmin_{g\in\Fcallambda} \|g - \regfunc\|_2^2.
\label{eqn:flOptFl}
\end{equation}
To prove this, consider any $g=\sumjM g^{(j)}\in\Fcallambda$.
Using the fact that $\Rcal(g) = \Rcal(\regfunc) + \|g-\regfunc\|_2^2$ for any $g$
and that $\|g\|_\Hcalaug \leq \Rlambda$ we obtain the above result as follows.
\begin{align*}
&\EE\big[(\regfunc(X)-Y)^2\big] + \|\funclambda - \regfunc\|_2^2 + \lambda\Rlambda^2
= \EE[(\funclambda(X) - Y)^2] + \lambda\Rlambda^2 \\
&\hspace{0.2in}\leq  \EE[(g(X)- Y)^2] + \lambda\sumjM \|g^{(j)}\|^2_\Hcalkj
\leq \EE\big[(\regfunc(X)-Y)^2\big] + \|g - \regfunc\|_2^2 + \lambda\Rlambda^2.
\end{align*}
By using the above, we get for all $\eta > 0$,
\begin{align*}
\EE\big[ \|\funchat - \regfunc\|_2^2 \big]
&\leq   (1+\eta) \EE\big[ \|\funclambda - \regfunc\|_2^2 \big] +
(1+1/\eta) \EE\big[\|\funchat - \funclambda\|_2^2 \big] \\
&= 
(1+\eta)\underbrace{\inf_{g\in\Fcallambda} \|g - \regfunc\|_2^2}_{\approxErr} +
  (1+1/\eta) \underbrace{\EE\big[\|\funchat - \funclambda\|_2^2 \big]}_{\estimErr}
\end{align*}
For the first step, by the AM-GM inequality we have
$2\int(\funchat-\funclambda)(\funclambda-\regfunc)
\leq 1/\eta \int (\funchat-\funclambda)^2 + \eta \int (\funclambda-\regfunc)^2$.
In the second step we have used~\eqref{eqn:flOptFl}.
The term $\approxErr$ is exactly as in Theorem~\ref{thm:agnosticThm} so we just need
to bound $\estimErr$. 

As before, we consider the RKHS $\Hcalaug$. Denote the representation of
$\funclambda$ in $\Hcalaug$ by $\funclambdavec =
(\funclambdaii{1},\dots,\funclambdaii{M})$.
Note that $\Rlambda = \|\funclambda\|_\Hcalaug$.
Analogous to the analysis in Appendix~\ref{sec:appMainThm} we define 
$\Deltaj = \funchatj - \funclambdaj$, $\Delta = \sumj\Deltaj = \funchat -
\funclambda$ and $\Deltavec = (\Deltaii{1},\dots,\Deltaii{M})$.
Note that $\estimErr = \EE[\|\Delta\|_2^2]$.

Let $\Deltaj = \sumlinf\deltalj\philj$ be the expansion of $\Deltaj$ in $L_2(\PX)$. 
For $t\in\NN$, which we will select later, define
$\Deltajdown = \sumlt\deltalj\philj$, $\Deltajup = \sumlgt\deltalj\philj$,
$\deltajdown = (\deltaii{1},\dots,\deltaii{t})\in\RR^t$ and
$\deltajup = (\deltalj)_{\ell>t}$.
Let $\Deltadown = \sumj\Deltajdown$ and $\Deltaup = \sumj\Deltajup$.
Continuing the analogy, let $\funclambdaj = \sumjM \thetalj\philj$ be the expansion
of $\funclambdaj$. Let $\thetajdown = (\thetaj_1,\dots,\thetaj_t) \in \RR^t$ and
$\thetadown = [\thetaiidown{1};\dots;\thetaiidown{M}] \in \RR^{tM}$.
Let $v\in\RR^n$ such that 
$\vj_i = \sumlgt\deltalj\philj(X_i)$ and $v_i = \sumj\vj_i$.
Let $\epsilon\in\RR^n$, $\epsilon_i = Y_i - \funclambda(X_i)$.
Also define the following quantities:
\[\sigmasqlambda(x) = \EE[(Y-\funclambda(X))^2|X=x], \hspace{0.3in}
\Blambdafour = 32\|\funclambdavec\|_\Hcalaug^4 + 8\taulambdafour/\lambda^2. \]
We begin with the following lemmas.
\\[\thmparaspacing]

\begin{lemma}
$\EE[\sigmafourlambda(X)] \leq 8\Psi^2\|\funclambdavec\|^4_\Hcalaug\rho^4 + 8\nu^4$.
\\[\thmparaspacing]
\label{lem:sigmaFourBound}
\end{lemma}

\begin{lemma}
$\EE\left[\big( \EE[\|\Deltavec\|^2_\Hcalaug | \Xn]\big)^2 \right] 
\leq \Blambdafour$.
\label{lem:BlambdaBound}
\end{lemma}

We first bound $\EE[\|\Deltajup\|_2^2] = \sumlgt\EE{\deltalj}^2$
using Lemma~\ref{lem:BlambdaBound} and Jensen's inequality.
\begin{equation}
\EE\left[\|\deltajup\|^2_2\right]
\;= \sumlgt\EE[{\deltalj}^2] 
\;\leq \muj_{t+1}\EE\left[\sumlgt\frac{{\deltalj}^2}{\mulj}\right]
\;\leq \muj_{t+1}\EE\left[\|\Deltaj\|^2_\Hcalkj\right]
\;\leq \muj_{t+1}\EE\left[\|\Deltavec\|^2_\Hcalaug\right]
\;\leq \muj_{t+1}\Blambdatwo
\label{eqn:agnosticDeltajup}
\end{equation}
Next we proceed to bound $\EE[\|\Deltadown\|^2_2]$. For this we will use 
$\Phij$,$\Philj$,$\Mj$,$\Mcal$, $Q$ from Appendix~\ref{sec:appMainThm}.
The first order optimality condition can be written as,
\[
\frac{1}{n}\sum_{i=1}^n \xi_{X_i} \left(\rinner{\xi_{X_i}}{\Deltavec} - \epsilon_i
\right) + \lambda \funchatvec = \zero.
\]
This has the same form as~\eqref{eqn:optCond} but the definitions of $\Deltavec$ and
$\epsilon_i$ have changed. Now, just as in the variance calculation, when we take the
$\Hcalaug$-inner product of the above with $(\zero,\dots,\philj,\dots,\zero)$ and
repeat for all $j$ we get,
\[
\left(I + Q^{-1}\left(\frac{1}{n}\Phi^\top \Phi - I\right) Q^{-1} \right)
Q\deltadown = -\lambda Q^{-1}\Mcal^{-1}\thetadown - \frac{1}{n}Q^{-1}\Phi^\top v
+ \frac{1}{n}Q^{-1}\Phi^\top \epsilon
\]
Since $\Phi$, $\Mcal$, $Q$ are the same as before we may reuse
Lemma~\ref{lem:technicalThree}. Then, as $Q\succeq I$ when the event $\Ecal$ holds,
\begin{align*}
\|\deltadown\|_2^2 \leq \|Q\deltadown\|_2^2 &\leq 4\|\lambda Q^{-1}\Mcal^{-1}\thetadown + 
\frac{1}{n}Q^{-1}\Phi^\top v + \frac{1}{n}Q^{-1}\Phi^\top \epsilon \|_2^2 \\
 &\leq 8 \|\frac{1}{n}Q^{-1}\Phi^\top v \|^2 + 
  8 \|\lambda Q^{-1}\Mcal^{-1}\thetadown - \frac{1}{n}Q^{-1}\Phi^\top \epsilon\|_2^2 
\numberthis \label{eqn:agnosticDeltadownBound}
\end{align*}
We now bound the two terms in the RHS in expectation via the following lemmas.
\\[\thmparaspacing]

\begin{lemma}
$ \EE[\|\frac{1}{n}Q^{-1}\Phi^\top v \|^2]
\leq \frac{1}{\lambda}M\Blambdatwo\rho^4\Psi\betat$
\\[\thmparaspacing]
\label{lem:technicalSix}
\end{lemma}

\begin{lemma}
$ \EE[\|\lambda Q^{-1}\Mcal^{-1}\thetadown - \frac{1}{n}Q^{-1}\Phi^\top
\epsilon\|_2^2] \leq \frac{1}{n}\rho^2 \gammakd(\lambda)\taulambdatwo $
\\[\thmparaspacing]
\label{lem:technicalSeven}
\end{lemma}

Now by Lemma~\ref{lem:BlambdaBound} we have,
$ \EE[\|\deltadown\|^2_2] = \PP(\Ecal) \EE[\|\deltadown\|_2^2|\Ecal] 
+ \EE[\indfone(\Ecal^c)\|\deltadown\|_2^2]
\leq \EE[\|\deltadown\|_2^2|\Ecal] + \Blambdatwo \PP(\Ecal^c)$.
$\EE[\|\deltadown\|_2^2|\Ecal]$ can be bounded using Lemmas~\ref{lem:technicalSix}
and~\ref{lem:technicalSeven} while $\PP(\Ecal^c)$ can be bounded using 
Lemma~\ref{lem:technicalThree}.
Combining these results along with~\eqref{eqn:agnosticDeltajup} 
we have the following bound for $\estimErr = \EE[\|\Delta\|_2^2]$,
\begin{align*}
\EE[\|\Delta\|_2^2] &\leq \EE\Bigg[\bigg\|\sumjM\Deltaj\bigg\|^2_2\Bigg] 
  \leq M\sumjM \EE\left[\|\Deltaj\|^2_2\right]
  = M\left( \EE[\|\deltadown\|_2^2] + \sumjM\EE[\|\deltajdown\|^2_2] \right)  \\
&\leq \frac{8}{n}M\rho^2 \gammakd(\lambda)\taulambdatwo +
\frac{8}{\lambda}M^2\Blambdatwo\rho^4\Psi\betat 
  + \Blambdatwo M\bigg(\frac{C\Md b(n,t,q)\rho^2\gammakd(\lambda)}{\sqrt{n}}\bigg)^q 
  + \Blambdatwo M\sumj\muj_{t+1}  
\end{align*}
Now we choose $t$ large enough so that the following are satisfied,
\[
\betat \leq \frac{\lambda}{M^2n\Blambdafour}\;,    \hspace{0.6in}
\sumjM\muj_{t+1} \leq \frac{1}{Mn\Blambdafour}\;, \hspace{0.6in}
\bigg(\frac{C\Md b(n,t,q)\rho^2\gammakd(\lambda)}{\sqrt{n}}\bigg)^q
\leq \frac{1}{Mn\Blambdafour}.
\]
Then the last three terms are $\bigO(1/n\Blambdatwo)$ 
and the first term dominates. Using Lemma~\ref{lem:sigmaFourBound} and recalling
that $\Rlambda^2 = \sumj{\Rlambdaj}^2 = \|\funclambdavec\|^2_\Hcalaug$ we get
$\estimErr \in \bigO\left(n^{-1} M \gammakd(\lambda) \Rlambda^2\right)$ 
as given in the theorem.

\vspace{0.2in}
\subsection*{Proofs of Technical Lemmas}

\subsection{Proof of Lemma~\ref{lem:BlambdaBound}}

Since $\funchat$ is the minimiser of the empirical objective,
\begin{align*}
\EE\left[ \lambda\|\funchatvec\|_\Hcalaug^2 | \Xn\right] &\leq
  \EE\left[ \lambda\sumjM\|\funchatj\|_\Hcalkj^2 + 
            \frac{1}{n}\sumin\left( \sumjM \funchatj(\Xj_i) - Y_i\right)^2 
            \;\Bigg| \Xn \right]\\
&\leq \EE\left[ \lambda\sumjM\|\funclambdaj\|_\Hcalkj^2 + 
            \frac{1}{n}\sumin\left( \sumjM \funclambdaj(\Xj_i) - Y_i\right)^2
            \;\Bigg| \Xn  \right]
\;\leq \lambda\|\funclambdavec\|^2_\Hcalaug + \frac{1}{n}\sumin \sigmasqlambda(X_i)
\end{align*}
Noting that $\Deltavec = \funchatvec -\funclambdavec$ and using the above bound and 
Jensen's inequality yields,
\[
\EE[\|\Deltavec\|^2_\Hcalaug | \Xn]
\;\leq 2\|\funclambdavec\|^2_\Hcalaug + 2\EE[\|\funchatvec\|^2_\Hcalaug|\Xn]
\;\leq 4\|\funclambdavec\|^2_\Hcalaug + \frac{2}{n\lambda}\sumin \sigmasqlambda(X_i)
\]
Applying Jensen's inequality once again yields,
\[
\EE[(\EE[\Deltavec\|^2_\Hcalaug|\Xn])^2]
\;\leq \EE\left[ \frac{8}{n^2\lambda^2}\left(\sumin\sigmasqlambda\right)^2 + 
  32\|\funclambdavec\|^4_\Hcalaug \right] 
\;\leq \frac{8}{n\lambda^2}\sumin\EE[\sigmafourlambda] + 
  32\|\funclambdavec\|^4_\Hcalaug
\;=\Blambdafour
\]

% \subsubsection{Proof of Lemma~\ref{lem:sigmaFourBound}}
\subsection{Proof of Lemma~\ref{lem:sigmaFourBound}}
First, using Jensen's inequality twice we have
\begin{align*}
\taulambdafour 
\;= \EE\left[\EE[(Y-\funclambda(X))^2|X]^2\right]
\;\leq \EE\left[(Y-\funclambda(X))^4\right]
\;\leq 8\EE[\funclambda^4(X)] + 8\EE[Y^4]
% \;\leq 8\EE[\funclambda^4(X)] + 8\nu^4
\numberthis \label{eqn:taulambdafirstbound}
\end{align*}
Consider any $\funclambdaj$,
\begingroup
\allowdisplaybreaks
\begin{align*}
\funclambdaj(x) 
&= \sumlinf \thetalj\philj(x) 
% = \sumlinf \big({\mulj}^{1/4}{\thetalj}^{1/2}\big)
% \left(\frac{{\thetalj}^{1/2}\philj(x)}{{\mulj}^{1/4}} \right)
\stackrel{(a)}{\leq} \left(\sumlinf{\mulj}^{1/3}{\thetalj}^{2/3}\right)^{3/4}
\left(\sumlinf\frac{{\thetalj}^{2}{\philj(x)}^4}{{\mulj}} \right)^{1/4} \\
&\stackrel{(b)}{\leq} 
\left(\sumjM\mulj\right)^{1/2}\left(\sumjM\frac{{\thetalj}^2}{\mulj}\right)^{1/4}
\left(\sumlinf\frac{{\thetalj}^{2}{\philj(x)}^4}{{\mulj}} \right)^{1/4} 
= {\Psij}^{1/2}\flambdajnorm^{1/2}
\left(\sumlinf\frac{{\thetalj}^{2}{\philj(x)}^4}{{\mulj}} \right)^{1/4} 
\end{align*}
\endgroup
In $(a)$, we used H\"older's inequality on ${\mulj}^{1/4}{\thetalj}^{1/2}$
and ${\thetalj}^{1/2}\philj(x)/{\mulj}^{1/4}$ with conjugates
$4/3$ and $4$ respectively.
In $(b)$ we used H\"older's inequality once again on ${\mulj}^{2/3}$ and
$({\thetalj}^2/{\mulj})^{1/3}$ with conjugates $3/2$ and $3$.
Now we expand $\funclambda$ in terms of the $\funclambdaj$'s  as follows,
\begin{align*}
\funclambda(x) &\leq
\sumjM {\Psij}^{1/2}\flambdajnorm^{1/2}
\left(\sumlinf\frac{{\thetalj}^{2}{\philj(x)}^4}{{\mulj}} \right)^{1/4} 
\leq \left(\sumjM\Psij\right)^{1/2}
\left(\sumjM\flambdajnorm 
\left(\sumlinf\frac{{\thetalj}^{2}{\philj(x)}^4}{{\mulj}} \right)^{1/2} \right)^{1/2}
\end{align*}
where we have applied Cauchy-Schwarz in the last step.
Using Cauchy-Schwarz once again,
\[
\funclambda^2(X) \leq \Psi \left(\sumjM\flambdajnorm^2\right)^{1/2}
  \left(\sumjM\sumlinf\frac{{\thetalj}^{2}{\philj(X)}^4}{{\mulj}} \right)^{1/2}
\]
Using Cauchy-Schwarz for one last time, we obtain
\[
\EE[\funclambda^4(x)] \;\leq \Psi^2 \flambdanormsq 
  \sumjM\sumlinf\frac{{\thetalj}^{2}{\EE[\philj(x)}]^4}{{\mulj}} 
\;\leq \Psi^2\flambdanormfour\rho^2
\]
where we have used Assumption~\ref{asm:basisTail} in the last step. When we combine
this with~\eqref{eqn:taulambdafirstbound} and use the fact that
$\EE[Y^4]\leq\nu^4$
 we get the statement of the lemma.

\subsection{Proof of Lemma~\ref{lem:technicalSix}}
The first part of the proof will mimic that of Lemma~\ref{lem:technicalFour}. By 
repeating the arguments for~\eqref{eqn:MPvIntermediate}, we get
\begin{align*}
\EE\left[\|\Mcal^{1/2}\Phi^\top v\|^2\right]
&\leq \sumjM\sumlt \mulj M \sumin \sumjpM
  \EE\left[ \EE[\|\Deltaii{j'}\|^2_\Hcalkii{j'}|\Xn] \|\Philj\|^2 
    \sumlpgt\mu^{(j')}_{\ell'} \phi^{(j')}_{\ell'}(X_i)^2
  \right] \\
&\leq M \sumjM\sumlt \sumin \sumjpM \sumlpgt
  \mulj \mu^{(j')}_{\ell'}  
    \EE\left[ \EE[\|\Deltaii{j'}\|^2_\Hcalkii{j'}|\Xn] \|\Philj\|^2 
    \phi^{(j')}_{\ell'}(X_i)^2
  \right] 
\end{align*}
Using Cauchy-Schwarz the inner expectation can be bounded via
$\sqrt{
\EE\left[\big(\EE[\|\Deltaii{j'}\|^2_\Hcalkii{j'}]\big)^2\right]
\EE\left[\|\Philj\|^4  \phi^{(j')}_{\ell'}(X_i)^4\right]
}$.
Lemma~\ref{lem:BlambdaBound} bounds the first expectation by $\Blambdafour$.
To bound the second expectation we use Assumption~\ref{asm:basisTail}.
\[
\EE\left[\|\Philj\|^4  \phi^{(j')}_{\ell'}(X_k)^4\right]
= \EE\left[ \Big(\sumin\philj(X_i)^2\Big)^2\philj(X_k)^4 \right] 
= \EE\left[ \sum_{i,i'}\philj(X_i)^2\philj(X_{i'})^2\philj(X_k)^4 \right] 
\leq n^2\rho^8
\]
Finally once again reusing some calculations from Lemma~\ref{lem:technicalFour},
\[
\EE\left[\left\|\frac{1}{n}Q^{-1}\Phi^\top v\right\|_2^2\right]
  \leq \EE\left[\frac{1}{\lambda}\left\| \frac{1}{n}M^{1/2}\Phi^\top v \right\|_2^2
          \right]
\leq \frac{M}{n^2\lambda}
  \underbrace{\left(\sumin  n\rho^4\right)}_{n^2\rho^4}
 \underbrace{\left(\sumjM\sumlt\mulj\right)}_{\Psi} 
  \underbrace{\left(\sumjpM \sumlpgt \mu^{(j')}_{\ell'} \right)}_{\betat}
\]

\subsection{Proof of Lemma~\ref{lem:technicalSeven}}
First note that we can write the LHS of the lemma as,
\[
\EE\left[ \left\|\lambda Q^{-1}\Mcal^{-1}\thetadown - \frac{1}{n}Q^{-1}\Phi^\top
\epsilon \right\|^2 \right]
=\sumjM\sumlt \frac{1}{1+\lambda/\mulj}
\EE\left[\left(\frac{\lambda\thetalj}{\mulj} - 
  \frac{1}{n}\sumin\philj(\Xj_i)\epsilon_i \right)^2\right]
\]
To bound the inner expectation we use the optimality conditions of the population
minimiser~\eqref{eqn:flambdaDefn}. We have,
\begin{align*}
2\EE\left[ \Big(\sumjM\funclambdaj(\Xj_i) - Y\Big)\xij_{X_i}\right] 
+ 2\lambda\funclambdaj =\zero
\;\;\implies\;\;
\EE\left[\xij_{X_i}\epsilon_i\right] = \lambda\funclambdaj
\;\;\implies\;\;
\EE\left[\philj(\Xj_i)\epsilon_i\right] = \lambda\frac{\thetalj}{\mulj}.
\numberthis \label{eqn:popOptCond}
\end{align*}
In the last step we have taken the $\Hcalaug$-inner product with
$(\zero,\dots,\philj,\dots,\zero)$.
Therefore the term inside the expectation is the variance of
$n^{-1}\sumi\philj(\Xj_i)\epsilon_i$ and can be bounded via,
\[
\VV\left[ \frac{1}{n} \sumin\philj(\Xj)\epsilon_i\right]
\;\leq \frac{1}{n}\EE\left[\philj(\Xj)^2\epsilon_i^2\right]
\;\leq \frac{1}{n}\sqrt{
\EE\left[ \philj(\Xj)^4\right] \EE\left[\epsilon_i^4\right]}
\;\leq\frac{1}{n} \rho^2 \taulambdatwo
\]
Hence the LHS can be bounded via,
\[
\frac{1}{n} \rho^2 \taulambdatwo \sumjM\sumlt \frac{1}{1+\lambda/\mulj}
\;\;=\;
\frac{1}{n} \rho^2 \gammakd(\lambda) \taulambdatwo 
\]

\vspace{0.2in}
\section{Some Details on Experimental Setup}
\label{sec:appExpDetails}

The function $f_d$ used in Figure~\ref{fig:knownorder} is the $\log$ of three
Gaussian bumps,
  \begin{align*}
  f_d(x) = \log\Bigg(
    \alpha_1\frac{1}{h_d^d} \exp\left( \frac{\|x-v_1\|^2}{2h_d^2}
\right) + 
  \hspace{0.1in}  \alpha_1\frac{1}{h_d^d} 
      \exp\left( \frac{\|x-v_2\|^2}{2h_d^2} \right) + 
    (1-\alpha_1-\alpha_2)\frac{1}{h_d^d} \exp\left( \frac{\|x-v_3\|^2}{2h_d^2}
\right)
    \Bigg) \numberthis
  \label{eqn:fdtilde}
  \end{align*}
where $h_d = 0.01\sqrt{d}$, $\alpha_1,\alpha_2\in[0,1]$ and $v_i\in\RR^d$ are
constant vectors.
For figures~\ref{fig:comporder}-\ref{fig:cvfour} we used $f_D$ where $D$ is 
given in the figures.
In all experiments, we used a test set of $2000$ points and plot the mean squared
test error.

For the real datasets, we normalised the training data so that the $X,y$ values
have zero mean and unit variance along each dimensions. 
We split the given dataset roughly equally to form a training
set and testing set. We tuned hyper-parameters via $5$-fold cross validation on the
training set and report the mean squared error on the test set.
For some datasets the test prediction error is larger than $1$. Such
datasets turned out to be quite noisy. In fact, when we used a constant predictor at
$0$ (i.e. the mean of the training instances) the mean squared error on the test set
was typically much larger than $1$. 

Below, we list details on the dataset: the source, the used predictor and features.

\newcommand{\datasetexpl}[5]{
  \textbf{#1:} (#2), \hspace{0.3in}Predictor: #3 \\
  Features: #4 
  #5}

\begin{enumerate}
\item \datasetexpl{Housing}{UCI}{CRIM}{All other attributes except CHAS which is
a binary feature.}{}

\item \datasetexpl{Galaxy}{SDSS data on Luminous Red Galaxies
from~\citet{tegmark06lrgs}}{Baryonic Density}{All other attributes.}{}

\item \datasetexpl{fMRI}{From~\cite{just10neurosemantic}}{Noun representation}{Voxel
Intensities. Since the actual dimensionality was very large, we use a random
projection to bring it down to 100 dimensions.}{}

\item \datasetexpl{Insulin}{From~\cite{tu14pancreatic}}{Insulin levels.}
{SNP features}{}

\item \datasetexpl{Skillcraft}{UCI}{TotalMapExplored}{All other attributes.}
{The usual predictor for this dataset is LeagueIndex but its an ordinal attribute and
not suitable for real valued prediction.}

\item \datasetexpl{School}{From Bristol Multilevel Modelling}{Given output}{Given
features.}{We don't know much about its
attributes. We used the given features and labels.}

\item \datasetexpl{CCPP*}{UCI}{Hourly energy output EP}{The other 4 features and 55
random features for the other 55 dimensions.}{}

\item \datasetexpl{Blog}{UCI Blog Feedback Dataset}{Number of comments in 24 hrs}
{The dataset had 280 features. The first 50 features were not used since they were
just summary statistics. 
Our features included features 51-62 given in the UCI website and the word counts
of 38 of the most frequently occurring words.
}{}

\item \datasetexpl{Bleeding}{From~\cite{guillame14utility}}{Given output}
{Given features reduced to 100 dimensions via a random projection.}
{We got this dataset from a private source and don't know much about its
attributes. We used the given features and labels.}

\item \datasetexpl{Speech}{Parkinson Speech dataset from UCI}{Median Pitch}
{All other attributes except the mean pitch, standard deviation, minimum pitch and
maximum pitches which are not actual features but statistics of the pitch.}{}

\item \datasetexpl{Music}{UCI}{Year of production}{All other attributes: 12 timbre
average and 78 timbre covariance}{}

\item \datasetexpl{Telemonit}{Parkinson's Telemonitoring dataset from UCI}
{total-UPDRS}{All other features except subject-id and motor-UPDRS (since it was too
correlated with total-UPDRS).}
{We only consider the female subjects in the dataset.}

\item \datasetexpl{Propulsion}{Naval Propulsion Plant dataset from UCI}
{Lever Position}{All other attributes.}
{We picked a random attribute as the predictor since no clear predictor was
specifified.}

\item \datasetexpl{Airfoil*}{Airfoil Self-Noise dataset from UCI}
{Sound Pressure Level}{The other 5 features and 35 random features.}{}

\item \datasetexpl{Forestfires}{UCI}{DC}{All other attributes.}
{We picked a random attribute as the predictor since no clear predictor was
specifified.}

  \item \datasetexpl{Brain}{From~\citet{wehbe14brain}}{Story feature at a
given time step}{Other attributes}{}

\end{enumerate}

\textbf{Some experimental details: }
\gps is the Bayesian interpretation of \krr. However, the results are
different in Table~\ref{tb:realData}. We believe this is due to differences in 
hyper-parameter tuning. For \gp, the 
GPML package~\cite{rasmussen06gps} optimises the
\gps marginal likelihood via L-BFGS. In contrast, our \krrs implementation
minimises the least squares cross validation error via grid search.
Some \addgps results are missing since it was very slow compared to 
other methods.
On the Blog dataset, \addkrrs took less than $35$s to train and all other
methods were completed in under 22 minutes. In contrast \addgps was not done training
even after several hours. Even on the relatively small speech dataset 
\addgps took about $80$ minutes.
Among the others, \backf, \mars, and \spams were the more expensive methods requiring
several minutes on datasets with large $D$ and $n$ 
whereas other methods took under 2-3 minutes.
We also experimented with locally cubic and quartic interpolation but exclude
them from the table since \locallin, \localquads generally performed better.
Appendix~\ref{sec:appExpDetails} has more details on the synthetic functions
and test sets.

{
\bibliography{kky}
}
\bibliographystyle{plainnat}

\end{document}